\documentclass[journal]{IEEEtran}

\ifCLASSINFOpdf
\else
\fi

\usepackage{graphicx}
\usepackage{subfigure}
\usepackage{multirow}
\usepackage{url}
\usepackage{makecell}
\usepackage{epstopdf}
\usepackage{mhchem}
\usepackage{epsfig}
\usepackage{amsfonts}
\usepackage{algpseudocode}
\usepackage{amsmath}
\usepackage[ruled, vlined, linesnumbered]{algorithm2e}

\makeatletter

\newcommand{\Rmnum}[1]{\expandafter\@slowromancap\romannumeral #1@}
\makeatother
\begin{document}
%
\title{Deep Learning for Spatio-Temporal Data Mining: A Survey}
%
%
%

\author{Senzhang Wang,
        Jiannong Cao,~\IEEEmembership{Fellow, IEEE},
        Philip S. Yu,~\IEEEmembership{Fellow, IEEE},
\thanks{S.Z. Wang (szwang@nuaa.edu.cn) is with the College of Computer Science and Technology, Nanjing University of Aeronautics and Astronautics, Nanjing, China, and also with the Department of Computing, The Hong Kong Polytechnic University, HongKong, China.}
\thanks{J.N. Cao is with the Department of Computing, The Hong Kong Polytechnic University, Hong Kong, China.}
\thanks{P.S. Yu is with the Department of Computer Science, University of Illinois at Chicago, Chicago, USA, and also with Institute for Data Science, Tsinghua University, Beijing, China.}
}


\maketitle

\begin{abstract}
With the fast development of various positioning techniques such as Global Position System (GPS), mobile devices and remote sensing, spatio-temporal data has become increasingly available nowadays. Mining valuable knowledge from spatio-temporal data is critically important to many real world applications including human mobility understanding, smart transportation, urban planning, public safety, health care and environmental management. As the number, volume and resolution of spatio-temporal datasets increase rapidly, traditional data mining methods, especially statistics based methods for dealing with such data are becoming overwhelmed. Recently, with the advances of deep learning techniques, deep leaning models such as convolutional neural network (CNN) and recurrent neural network (RNN) have enjoyed considerable success in various machine learning tasks due to their powerful hierarchical feature learning ability in both spatial and temporal domains, and have been widely applied in various spatio-temporal data mining (STDM) tasks such as predictive learning, representation learning, anomaly detection and classification. In this paper, we provide a comprehensive survey on recent progress in applying deep learning techniques for STDM. We first categorize the types of spatio-temporal data and briefly introduce the popular deep learning models that are used in STDM. Then a framework is introduced to show a general pipeline of the utilization of deep learning models for STDM. Next we classify existing literatures based on the types of ST data, the data mining tasks, and the deep learning models, followed by the applications of deep learning for STDM in different domains including transportation, climate science, human mobility, location based social network, crime analysis, and neuroscience. Finally, we conclude the limitations of current research and point out future research directions.
\end{abstract}


\IEEEpeerreviewmaketitle

\section{Introduction}
Spatio-temporal data mining (STDM) is becoming growingly important in the big data era with the increasing availability and importance of large spatio-temporal datasets such as maps, virtual globes, remote-sensing images, the decennial census and GPS trajectories. STDM has broad applications in various domains including environment and climate (e.g. wind prediction and precipitation forecasting), public safety (e.g. crime prediction), intelligent transportation (e.g. traffic flow prediction), human mobility (e.g. human trajectory pattern mining), etc. Classical data mining techniques that are used to deal with transaction data or graph data often perform poorly when applied to spatio-temporal datasets because of many reasons. First, ST data are usually embedded in continuous space, whereas classical datasets such as transactions and graphs are often discrete. Second, patterns of ST data usually present both spatial and temporal properties, which is more complex and the data correlations are hard to capture by traditional methods. Finally, one of the common assumptions in traditional statistical based data mining methods is that data samples are independently generated. When it comes to the analysis of spatio-temporal data, however, the assumption about the independence of samples usually does not hold because ST data tends to be highly self correlated. 

Although STDM has been widely studied in the last several decades, a common issue is that traditional methods largely rely on feature engineering. In other words, conventional machine learning and data mining techniques for STDM are limited in their ability to process natural ST data in their raw form. For example, to analyze human's brain activity from fMRI data, usually careful feature engineering and considerable domain expertise are needed to design a feature extractor that transforms the raw data (e.g. the pixel values of the scanned fMRI images) into a suitable internal representation or feature vector. Recently, with the prevalence of deep learning, various deep leaning models such as convolutional neural network (CNN) and recurrent neural network (RNN) have enjoyed considerable success in various machine learning tasks due to their powerful hierarchical feature learning ability, and have been widely applied in many areas including computer vision, natural language processing, recommendation, time series data prediction, and STDM. Compared with traditional methods, the advantages of deep learning models for STDM are as follows.

\begin{itemize}
\item \textbf{Automatic feature representation learning} Significantly different from traditional machine learning methods that require hand-crafted features, deep learning models can automatically learn hierarchical feature representations from the raw ST data. In STDM, the spatial proximity and the long-term temporal correlations of the data are usually complex and hard to be captured. With the multi-layer convolution operation in CNN and the recurrent structure of RNN, such spatial proximity and temporal correlations in ST data can be automatically and effectively learned from the raw data directly.

\item \textbf{Powerful function approximation ability} Theoretically, deep learning can approximate any complex non-linear functions and can fit any curves as long as it has enough layers and neuros. Deep learning models usually consist of multiple layers, and each layer can be considered as a simple but non-linear module with pooling, dropout, and activation functions so that it transforms the feature representation at one level into a representation at a higher and more abstract level. With the composition of enough such transformations, very complex functions can be learned to perform more difficult STDM tasks with more complex ST data.

\end{itemize}

\begin{figure}[!t]
\begin{center}
\includegraphics[height=4.8cm]{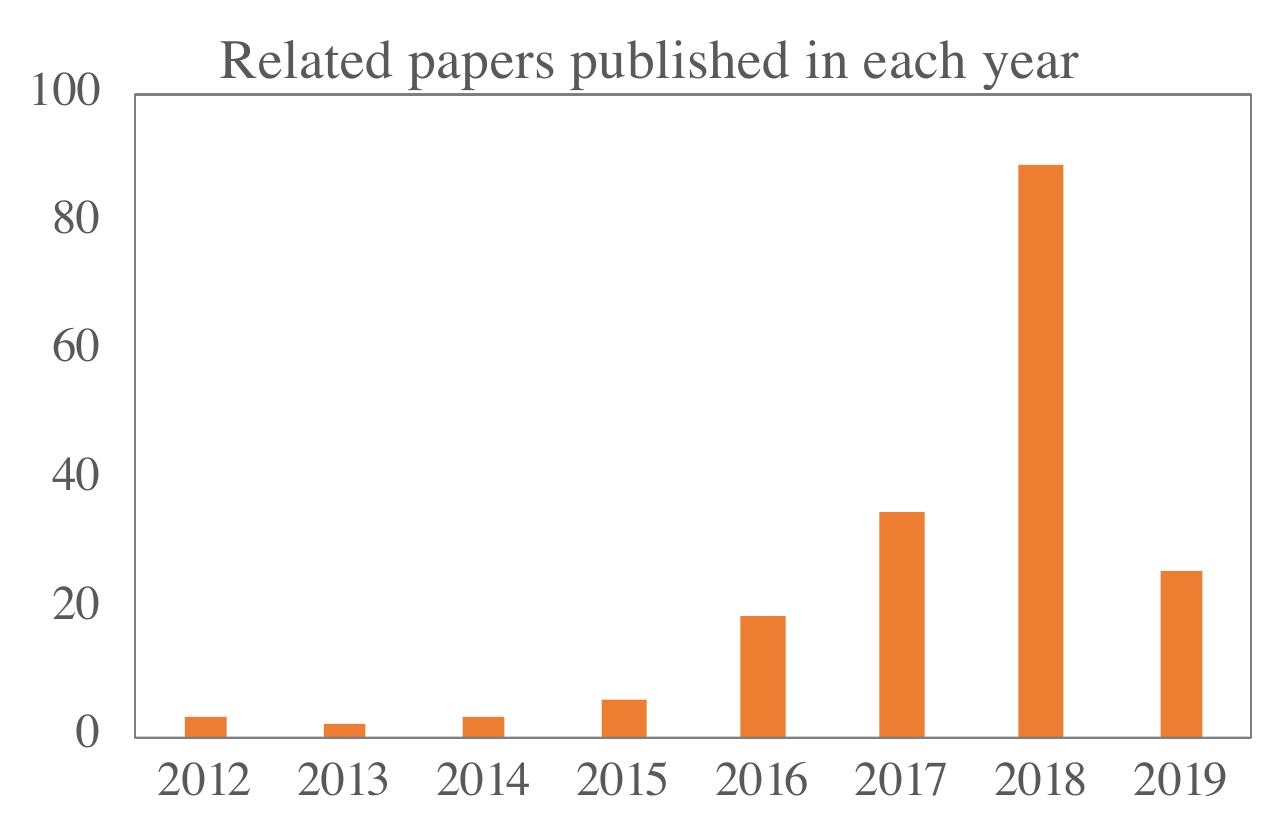}
\caption{Number of papers that explore deep learning techniques for STDM published in recent years.}
\label{num}
\end{center}
\end{figure}

Figure \ref{num} shows the number of yearly published papers that explore deep learning techniques for various STDM tasks. One can see that there is a significant increase trend of the paper number in the last three years. Only less than 10 related papers are published each year from 2012 to 2015. From 2016 on, the number increases rapidly and many researchers try different deep learning models for different types of ST data in different applications domains. In 2018, there are about 90 related papers published, which is a large number. The complete number for 2019 is unavailable now, but we believe that the growing trend will keep on this year and also the future several years. Given the richness of problems and the variety of real applications, there is an urgent need for overviewing the existing works that explore deep learning techniques in the rapidly advancing field of STDM due to the following reasons. It can highlight the similarities and differences of using different deep learning models for addressing STDM problems of diverse applications domains. This can enable the cross-pollination of ideas across disparate research areas and application domains, by making it possible to see how deep learning model (e.g., CNN and RNN) developed for a certain problem in a particular domain (e.g., traffic flow prediction in transportation) can be useful for solving a different problem in another domain (e.g., crime prediction in crime analysis).

\textbf{Related surveys on STDM} There are a few recent surveys that have reviewed the literature on STDM in certain contexts from different perspectives. \cite{STDM15} and \cite{BigSpatial} discussed the computational issues for STDM algorithms in the era of ``big data'' for application domains such as remote sensing, climate science, and social media analysis. \cite{pattern} focused on frequent pattern mining from spatio-temporal data. It stated the challenges of pattern discovery from ST data and classified the patterns into three categories: individual periodic pattern; pairwise movement pattern and aggregative patterns over multiple trajectories. \cite{handbook} reviewed the state-of-the-art in STDM research and applications, with emphasis placed on the data mining tasks of prediction, clustering and visualization for spatio-temporal data. \cite{Geo-Infor} reviewed STDM from the computational perspective, and emphasized the statistical foundations of STDM. \cite{trajectory} reviewed the methods and applications for trajectory data mining, which is an important type of ST data. \cite{STCluster} provided a comprehensive survey on ST data clustering. \cite{atluri2018spatio} discussed different types of ST data and the relevant data mining questions that arise in the context of analyzing each type of data. They classify literature on STDM into six major categories: clustering, predictive learning, change detection, frequent pattern mining, anomaly detection, and relationship mining. However, all these works reviewed STDM from the perspective of traditional methods rather than deep learning methods.  \cite{nguyen2018deep} and \cite{wang2018enhancing} provided a survey that specially focused on the utilization of deep learning models for analyzing traffic data to enhance the intelligence level of transportation systems. There still lacks of a broad and systematic survey on exploring deep learning techniques for STDM in general.

\textbf{Our Contributions} Compared with existing works, our paper makes notable contributions summarized as follows:
\begin{itemize}
\item \textbf{First survey} To our knowledge, this is the first survey that reviews recent works exploring deep learning techniques for STDM. In light of the increasing number of studies on deep learning for spatio-temporal data analytics in the last several years, we first categorize spatio-temporal data types, and present the popular deep learning models that are widely used in STDM. We also summarize the data representations for different data types, and summarize which deep learning models are suitable to handle which types of data representations of ST data.

\item \textbf{General framework} We present a general framework for deep learning based STDM, which consists of the following major steps: data instance construction, data representation, deep learning model selection and addressing STDM problems. Under the guidance of the framework, given a particular STDM task, one can better use the proper data representations and select or design the suitable deep learning models for the task under study.

\item \textbf{Comprehensive survey} This survey provides a comprehensive overview on recent advances of using deep learning techniques for different STDM problems including predictive learning, representation learning, classification, estimation and inference, anomaly detection, and others. For each task, we provide detailed descriptions on the representative works and models for different types of ST data, and make necessary comparison and discussion. We also categorize and summarize current works based on the application domains including transportation, climate science, human mobility, location based social network, crime analysis, and neuroscience.

\item \textbf{Future research directions} This survey also highlights several open problems that are not well studied currently, and points out possible research directions in the future.
\end{itemize}

\textbf{Organization of This Survey} The rest of this survey is organized as follows. Section II introduces the categorization of the ST data, and briefly introduces the deep learning models that are widely used for STDM. Section III provides
a general framework for using deep learning for STDM. Section IV overviews various STDM tasks addressed by deep learning models. Section V
presents a gallery of applications across various domains. Section VI discusses the limitations of existing works and suggests future directions. We finally conclude this paper in Section VII.

\section{Categorization of Spatio-Temporal Data}
\subsection{Data Types}
There are various types of ST data that differs in the way of data collection and representation in different real-world applications. Different application scenarios and ST data types lead to different categories of data mining tasks and problem formulations. Different deep learning models usually have different preferences to the types of ST data and have different requirements for the input data format. For example, CNN model is designed to process image-like data, while RNN is usually used to process sequential data. Thus it is important to first summarize the general types of ST data and represent them properly. 
We follow and extend the categorization in \cite{atluri2018spatio}, and classify the ST data into the following types: event data, trajectory data, point reference data, raster data, and videos. 

\begin{figure}
\centering 
\subfigure[Three types of events]{
\label{Fig.sub.1}
\includegraphics[width=0.21\textwidth]{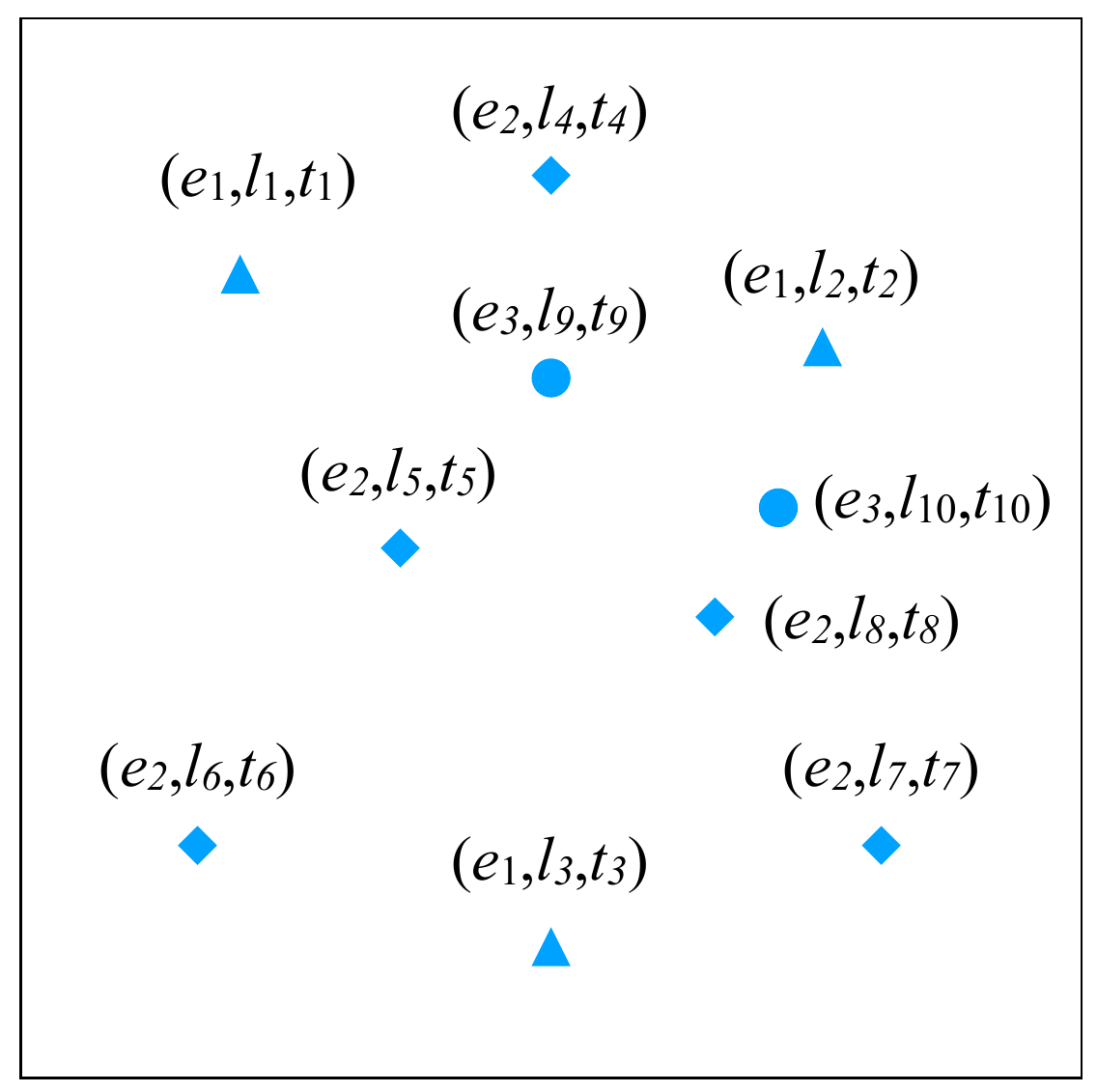}}
\subfigure[The trajectories of two moving objects]{
\label{Fig.sub.2}
\includegraphics[width=0.21\textwidth]{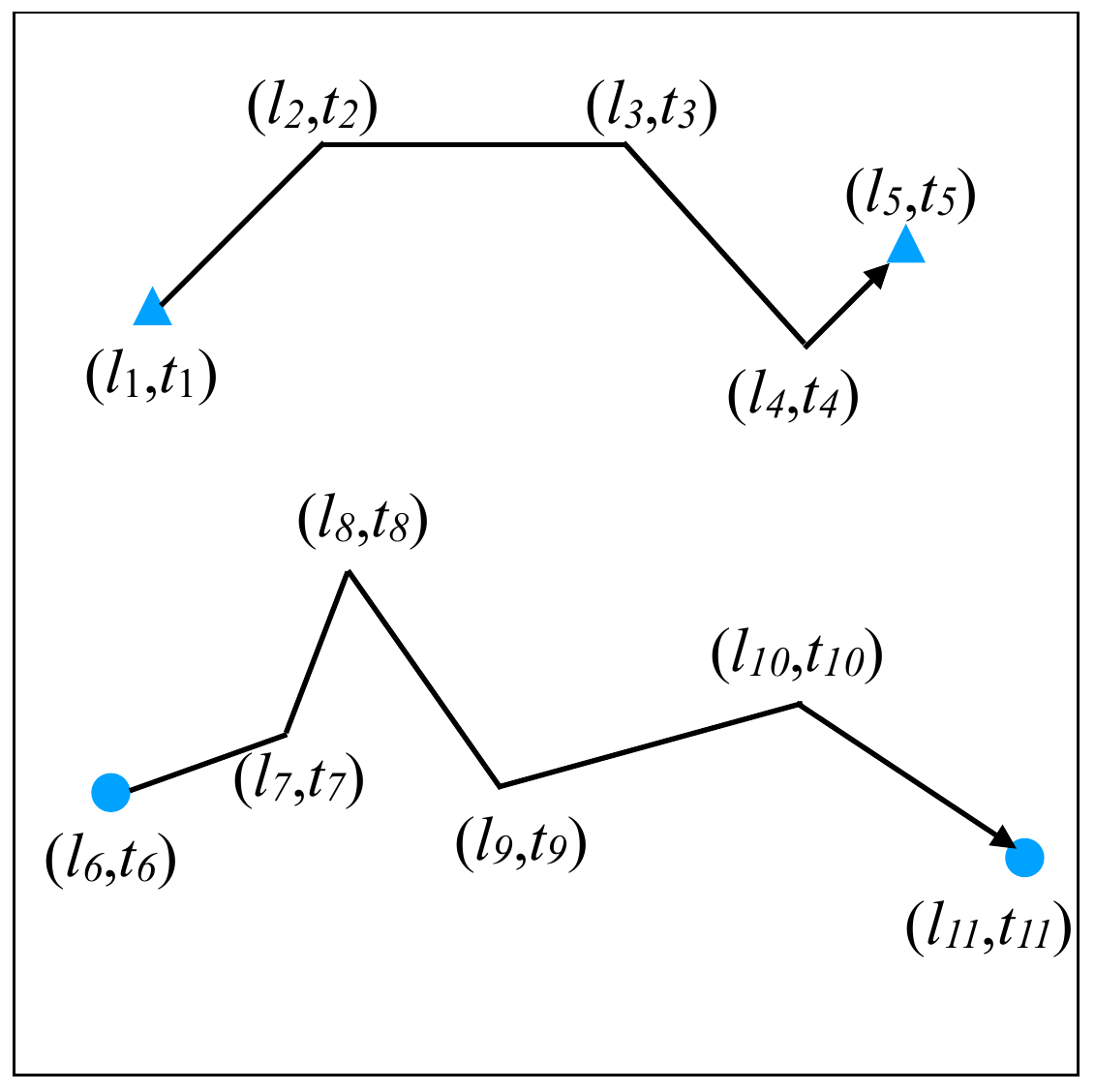}}
\caption{Illustration of event and trajectory data types}
\label{Fig.main}
\end{figure}
\textbf{Event data.} Event data comprise of discrete events occurring at point locations and times (e.g., crime
events in the city and traffic accident events in a transportation network). An event can generally be characterized by a point location and time,
which denotes where and when the event occurred, respectively. For example, a crime event can be characterized as such a tuple $(e_i, l_i, t_i)$, where $e_i$ is the crime type, $l_i$ is the location where the crime occurs and $t_i$ is the time when it occurs. Fig. 1(a) shows an illustration of the event data. It shows three types of events denoted by different shapes of the symbol. ST event data are common in real-world applications such as criminology (incidence of crime and related events), epidemiology (disease outbreak events), transportation (car accident), and social network (social event and trending topics). 

\textbf{Trajectory data.} Trajectories denote the paths traced by bodies moving in space over time. (e.g., the moving route of a bike trip or taxi trip). Trajectory data are usually collected by the sensors deployed on the moving objects that can periodically transmit the location of the object over time, such as GPS on a taxi. Fig. 1(b) shows an illustration of two trajectories. Each trajectory can be usually characterized as such a sequence $\{(l_1,t_1), (l_2,t_2)...(l_n, t_n)\}$, where $l_i$ is the location (e.g. latitude and longitude) and $t_i$ is the time when the moving object passes this location. Trajectory data such as human trajectory, urban traffic trajectory and location based social networks are becoming ubiquitous with the development of Mobile applications and IoT techniques. 

\begin{figure}[!t]
\begin{center}
\includegraphics[height=3.5cm]{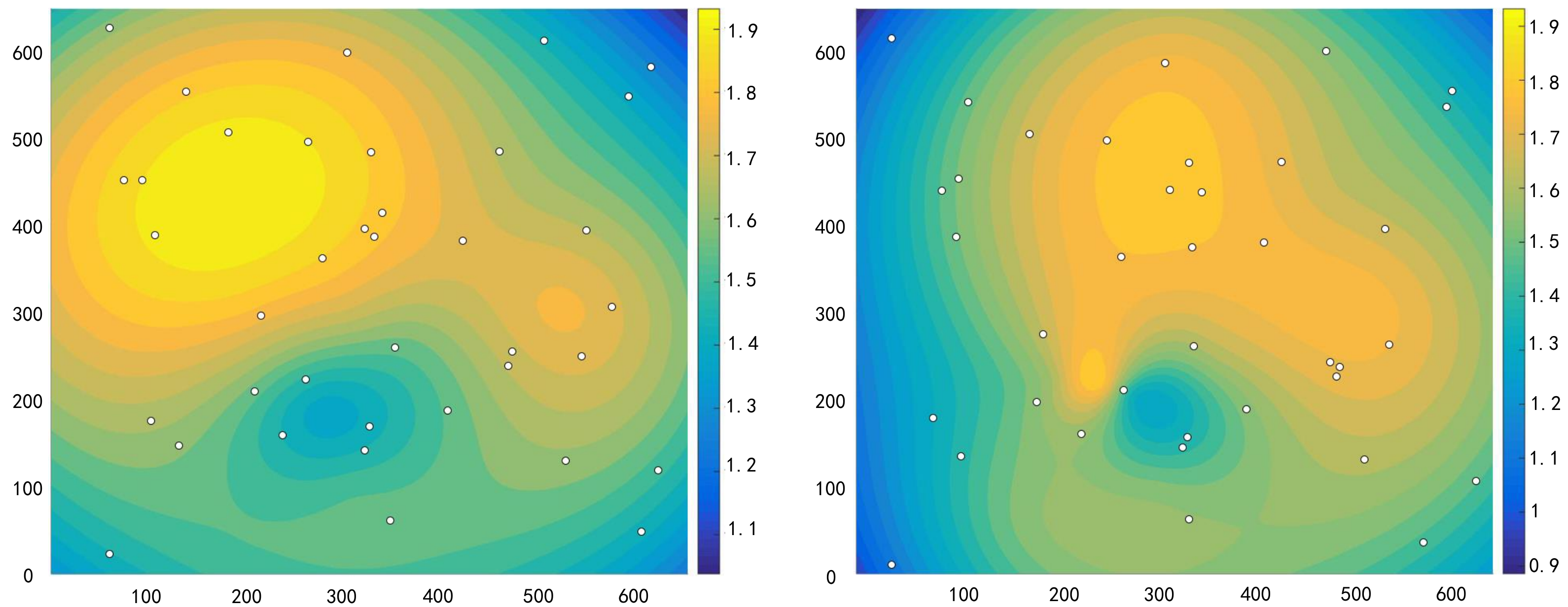}
\caption{Illustration of ST reference point data in two time stamps. The while circles are the locations of the sensors that record the readings of the ST field. The color bars show the distribution of the ST field.}
\label{STref}
\end{center}
\end{figure}

\textbf{Point reference data.} Point reference data consist of measurements of a continuous ST field such as temperature, vegetation, or population over a set of moving reference points in space and time. For example, meteorological data such as temperature and humidity are commonly measured using weather balloons floating in space, which continuously record weather observations. Point reference data can be usually represented as a set of tuples as follows $\{(r_1,l_1,t_1),(r_2,l_2,t_2)...(r_n,l_n,t_n)\}$. Each tuple $(r_i, l_i, t_i)$ denotes the measurement of a sensor $r_i$ at the location $l_i$ of the ST filed at time $t_i$. Fig. \ref{STref} shows an example of the point reference data (e.g. sea surface temperature) in a continuous ST field at two time stamps. They are measured by the sensors at reference locations (shown as while circles) on the two time stamps. Note that the locations of the temperature sensors change over time.

\begin{figure}[!t]
\begin{center}
\includegraphics[height=4.4cm]{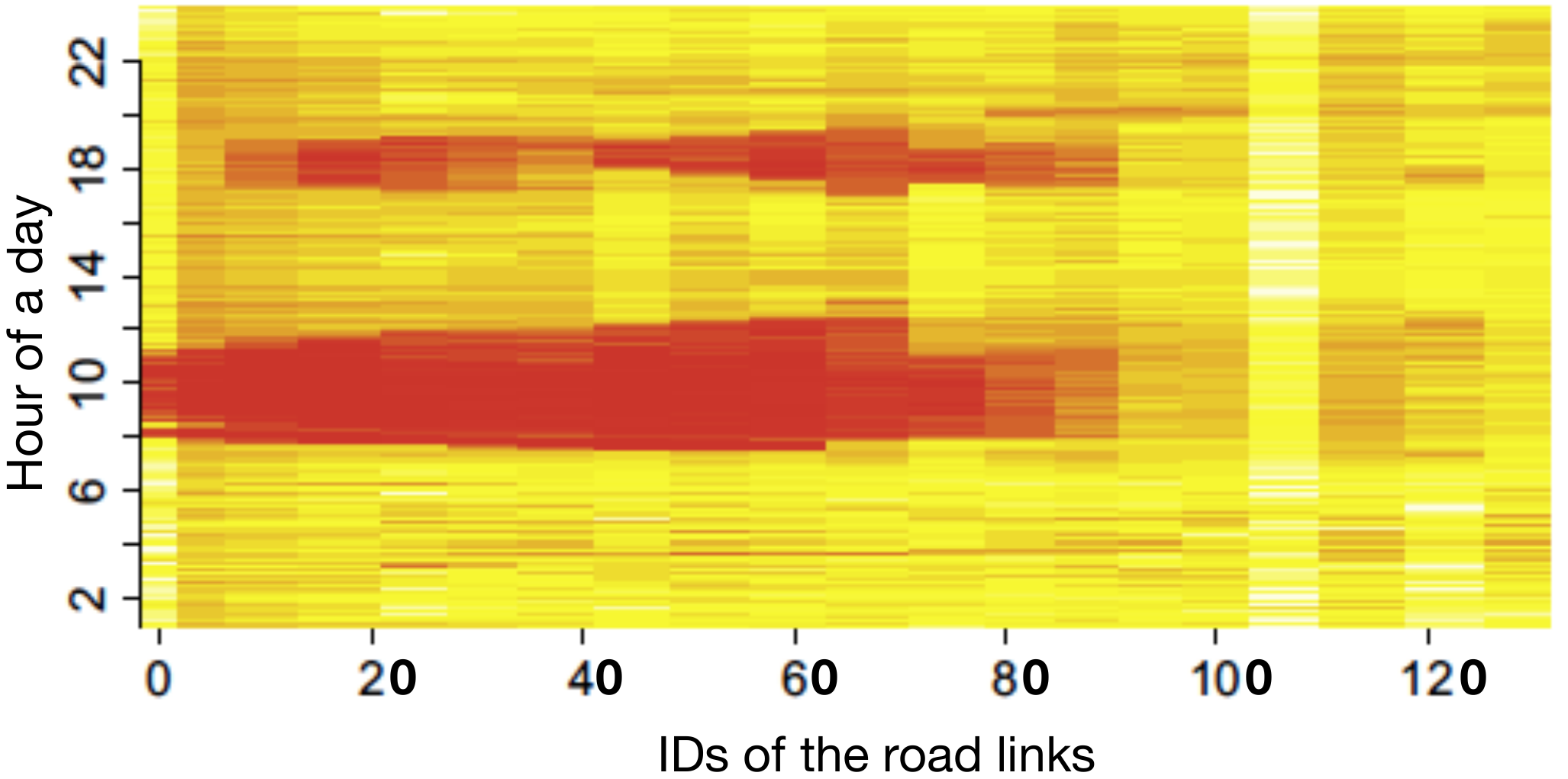}
\caption{Illustration of raster data collected from traffic flow sensors. The x-axis is the ID of the road links in a transportation network, and the y-axis is the hour of a day. Different colors denote different traffic flows on the road links captured by the road sensors deployed at fixed locations.}
\label{raster}
\end{center}
\end{figure}

\textbf{Raster data.} Raster data are the measurements of a continuous or discrete ST field that are recorded at fixed locations in space and at fixed time points. The major difference between point reference data and raster data is that the locations of the point reference data keep changing while the locations of the raster data are fixed. The locations and times for measuring the ST field can be regularly or irregularly distributed. Given $m$ fixed locations $S=\{s_1,s_2,...s_m\}$ and $n$ time stamps $T=\{t_1,t_2,...t_n\}$, the raster data can be represented as a matrix $R^{m\times n}$, where each entry $r_{ij}$ is the measurement at location $s_i$ at time stamp $t_j$. Raster data are also quite common in real-world applications such as transportation, climate science, and neuroscience. For example, the air quality data (e.g. PM2.5) can be collected by the sensors deployed at fixed locations of a city, and the data collected in a continuous time period form the air quality raster data. In neuroscience, functional magnetic resonance imaging or functional MRI (fMRI) measures brain activity by detecting changes associated with blood flow. The scanned fMRI signals also form the raster data for analyzing the brain activity and identifying some diseases. Fig. \ref{raster} shows an example of the traffic flow raster data of a transportation network. Each road is deployed a traffic sensor to collect real time traffic flow data. The traffic flow data of all the road sensors in a whole day (24 hours) form a raster data.

\textbf{Video.} A video that consists of a sequence of images can be also considered as a type of ST data. In the spatial domain, the neighbor pixels usually have similar RGB values and thus present high spatial correlations. In the temporal domain, the images of consecutive frames usually change smoothly and present high temporal dependency. A video can be generally represented as a three dimensional tensor with one dimension representing time $t$ and the other two representing an image. Actually, video data can be also considered as a special raster data if we assume that there is a ``sensor'' deployed at each pixel and at each frame the ``sensors'' will collect the RGB values. Deep learning based video data analysis is extremely hot and a large number of papers are published in recent years. Although we categorize videos as a type of ST data, we focus on reviewing related works from the perspective of data mining and video data analysis falls into the research areas of computer vision and pattern recognition. Thus in this survey we do not cover the ST data type of videos. 



\subsection{Data Instances and Representations}
The basic unit of data that a data mining algorithm operates upon is called a data instance. For a classical data mining setting, a data instance can be usually represented as a set of features with a label for supervised learning or without labels for unsupervised learning. In the ST data mining scenario, there are different types of data instances for different ST data types. For different data instances, there are several types of data representations that are used to formulate the data for further mining by the deep learning models.

\begin{figure}[!t]
\begin{center}
\includegraphics[height=5cm]{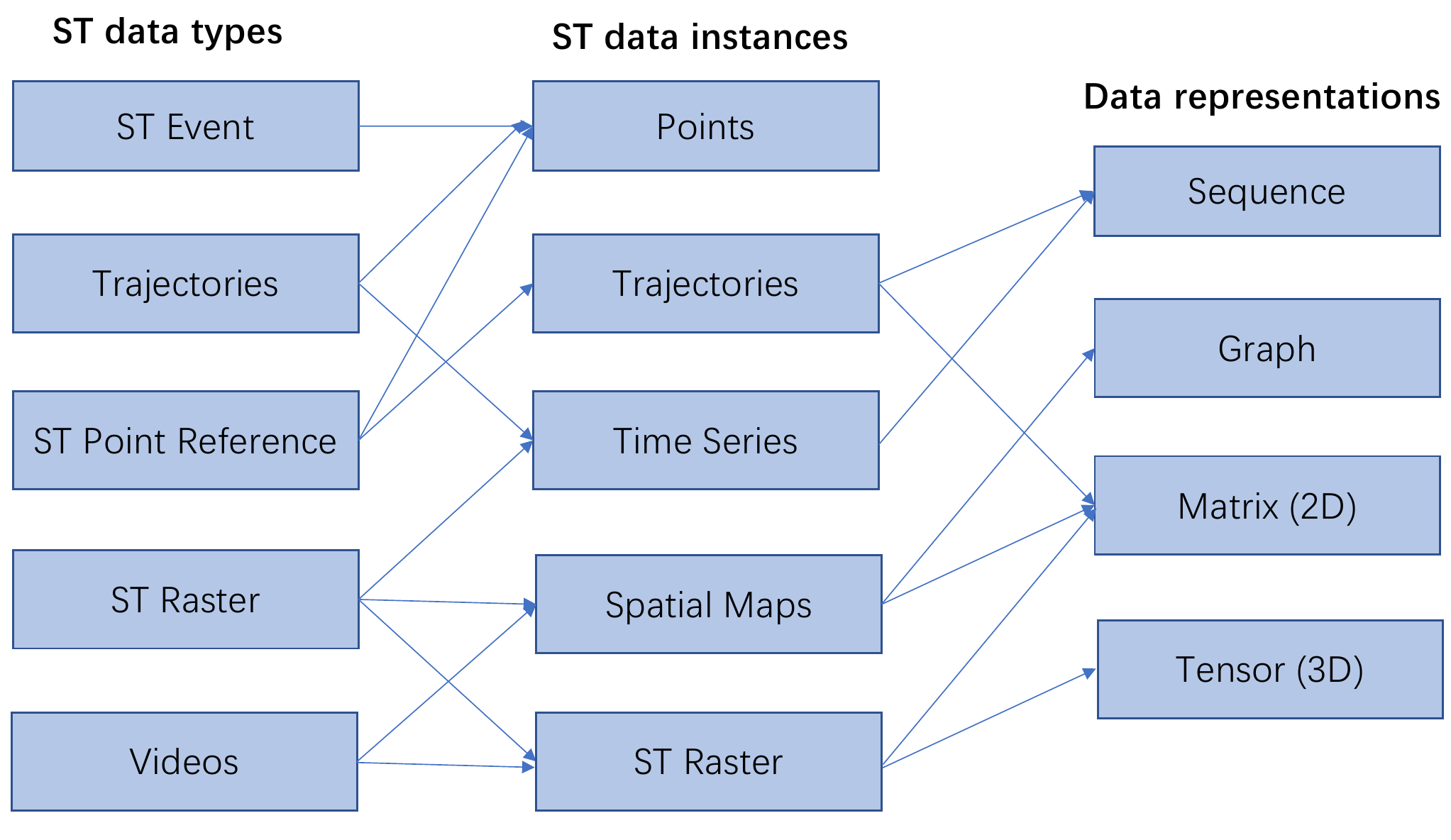}
\caption{Data instances and representations of different ST data types}
\label{datarepre}
\end{center}
\end{figure}

\textbf{Data instances.} In general, the ST data can be summarized into the following data instances: points, trajectories, time series, spatial maps and ST raster as shown in the left part of Fig. \ref{datarepre}. A ST point can be represented as a tuple containing the spatial and temporal information as well as some additional features of an observation such as the types of crimes or traffic accidents.  Besides ST events, trajectories and ST point reference can also be formed as points. For example, one can break a trajectory into several discrete points to count how many trajectories have passed a particular region in a particular time slot. Besides formed as points and trajectories, trajectories can be also formed as time series in some applications. If we fix the location and count the number of trajectories traversing the location, it forms a time series data. The data instance of spatial maps contains the data observations of all the sensors in the entire ST filed at each time stamp. For example, the traffic speed readings of all the loop sensors deployed at the expressway at time $t$ form a spatial map data. The data instance of the ST raster data contains the measurements spanning the entire set of locations and time stamps. That is, a ST raster comprises of a set of spatial maps. 

Different data instances can be extracted from ST raster as time series, spatial maps or ST raster itself, depending on different applications and analytic requirements. First, we can consider the measurements at a particular ST grid of the ST field as a time series for some time series mining tasks. Second, for each time stamp the measurements of an ST raster can be considered as a spatial map. Third, one can also consider all the measurements spanning all the locations and time stamps as a whole for analysis. In such a case, ST raster itself can be a data instance. 

\textbf{Data representations.} For the above mentioned five types of ST data instances, four types of data representations are generally utilized to represent them as the input of various deep learning models, sequence, graph, 2-dimensional matrix and 3-dimensional tensor as shown in the right part of Fig. \ref{datarepre}. Different deep learning models require different types of data representations as input. Thus how to represent the ST data instances relies on the data mining task under study and the selected deep learning model.

Trajectories and time series can be both represented as sequences. Note that trajectories sometime are also represented as a matrix whose two dimensions are the row and column ids of grid ST field. Each entry value of the matrix denotes whether the trajectory traverses the corresponding grid region. Such a data representation is usually used to facilitate the utilization of CNN models \cite{karatzoglou2018convolutional,ouyang2016deepspace,varshneya2017human}. Although graph can be also represented as a matrix, here we categorize graph and image matrix as two different types of data representations. This is because graph nodes does not follow the Euclidean distance as the image matrix does, and thus the way to deal with graphs and image matrices are totally different. We will discuss more details on the methods to handle the two types of data representations later. Spatial maps can be both represented as graphs and matrices, depending on different applications. For example, in urban traffic flow prediction, the traffic data of a urban transportation network can be represented as a traffic flow graph \cite{li2018diffusion,wang2018efficient} or cell region-level traffic flow matrix \cite{polson2017deep,sun2017dxnat}. Raster data are usually represented as 2D matrices or 3D tensors. For the case of matrix, the two dimensions are locations and time steps, and for the case of tensor, the three dimensions are row region cell id, column region id and time stamp. Matrix is a simpler data representation format compared with tensor, but it loses the spatial correlation information among the locations. Both are widely used to represent raster data. For example, in wind forecasting, the wind speed time series data of multiple anemometers deployed in different locations are usually merged as a matrix, and then is feed into a CNN or RNN model for future wind speed prediction \cite{liu2018smart,zhu2018wind}. In neuroscience, one's fMRI data are a sequence of scanned fMRI brain images, and thus can be represented as a tensor like a video. Many works use the fMRI images tensor as the input of CNN model for feature learning to detect the brain activity \cite{jin2014classification,kleesiek2016deep} and diagnose diseases \cite{nie20163d,wen2018deep}.

\subsection{Preliminary of Deep Learning Models}
In this subsection, we briefly introduce several deep learning models that are widely used for STDM, including RBM, CNN, GraphCNN, RNN, LSTM, AE/SAE, and Seq2Seq.
\begin{figure}[t]
\begin{center}
\includegraphics[height=4cm]{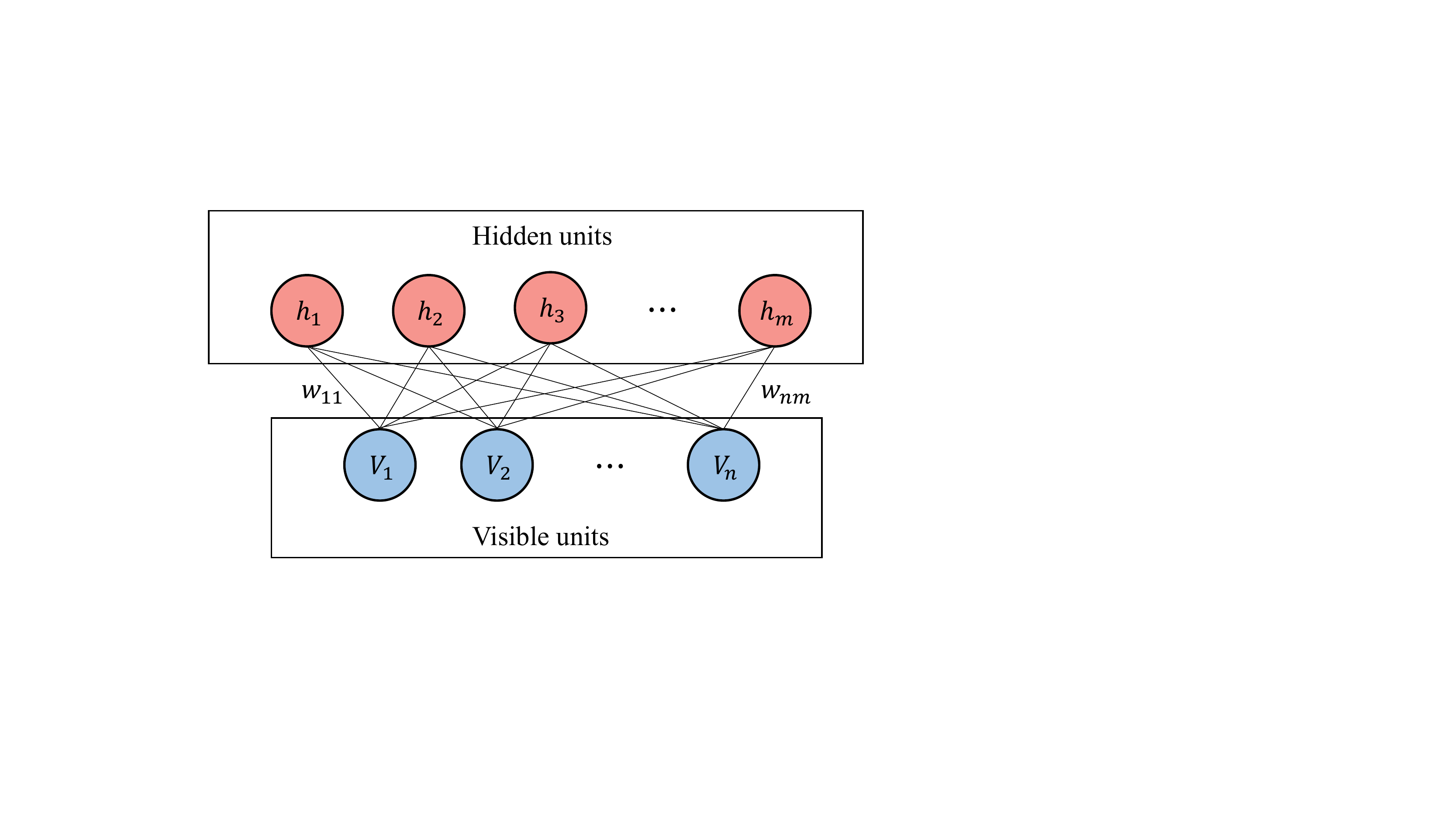}
\caption{Structure of the RBM model.}
\label{rbm}
\end{center}
\end{figure}

\textbf{Restricted Boltzmann Machines (RBM).} A Restricted Boltzmann Machine is a two-layer stochastic neural network \cite{RBM} which can be used for dimensionality reduction, classification, feature learning and collaborative filtering. As shown in Fig. \ref{rbm}, the first layer of the RBM is called the visible, or input layer with the neuron nodes $\{v_1,v_2,...v_n\}$, and the second is the hidden layer with the neuron nodes $\{h_1,h_2,...h_m\}$. As a fully-connected bipartite undirected graph, all nodes in RBM are connected to each other across layers by undirected weight edges $\{w_{11},...w_{nm}\}$, but no two nodes of the same layer are linked. The standard type of RBM has a binary-valued nodes and also bias weights. RBM tries to learn a binary code or representation of the input, and depending on the particular task, RBM can be trained in either supervised or unsupervised ways. RBM is usually used for learning features.

\begin{figure}[t]
\begin{center}
\includegraphics[height=4.5cm]{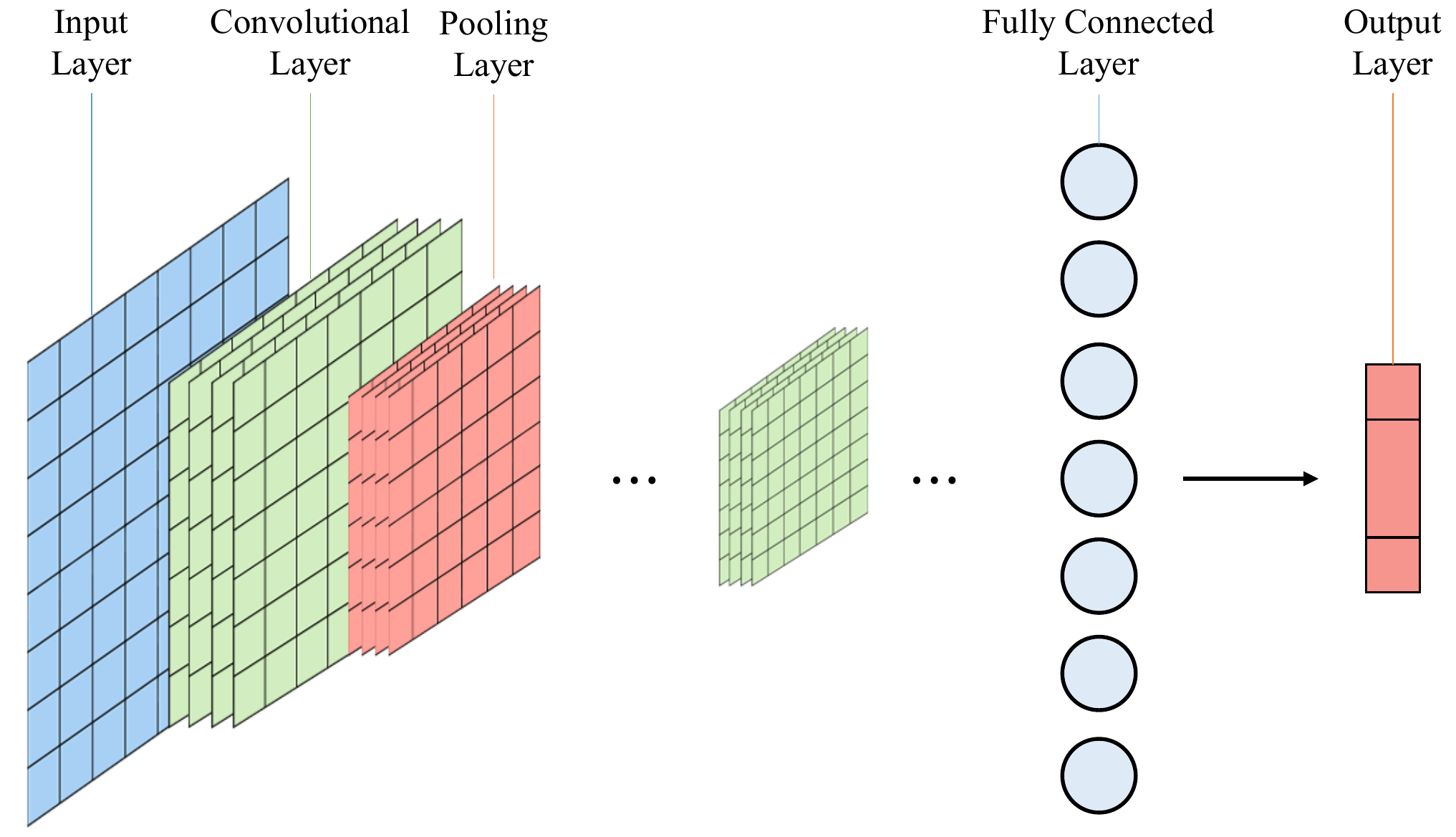}
\caption{Structure of the CNN model.}
\label{cnn}
\end{center}
\end{figure}
\begin{figure}[!t]
\begin{center}
\includegraphics[height=4.5cm]{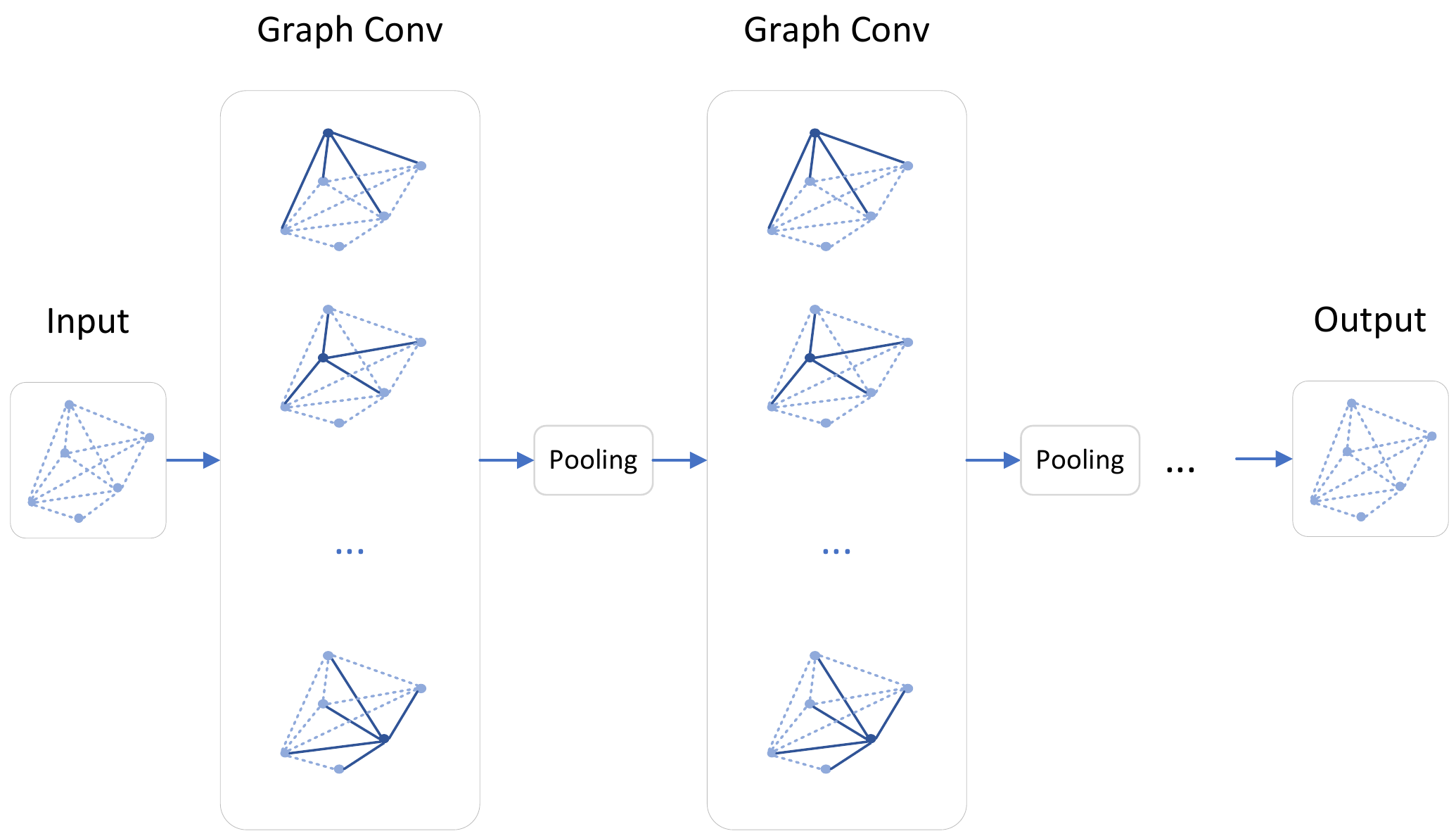}
\caption{Structure of GraphCNN model.}
\label{GCN}
\end{center}
\end{figure}
\textbf{CNN.} Convolutional neural networks (CNN) is a class of deep, feed-forward artificial neural networks that are applied to analyze visual imagery. A typical CNN model usually contains the following layers as shown in Fig. \ref{cnn}: the input layer, the convolutional layer, the pooling layer, the fully-connected layer and the output layer. The convolutional layer will determine the output of neurons of which are connected to local regions of the input through the calculation of the scalar product between their weights and the region connected to the input volume. The pooling layer will then simply perform downsampling along the spatial dimensionality of the given input to reduce the number of parameters. The fully-connected layers will connect every neuron in one layer to every neuron in the next layer to learn the final feature vectors for classification. It is in principle the same as the traditional multi-layer perceptron neural network (MLP). Compared with traditional MLPs, CNNs have the following distinguishing features that make them achieve much generalization on vision problems: 3D volumes of neurons, local connectivity and shared weights. CNN is designed to process image data. Due to its powerful ability in capturing the correlations in the spatial domain, it is now widely used in mining ST data, especially the spatial maps and ST rasters.

\textbf{GraphCNN.} CNN is designed to process images which can be represented as a regular grid in the Euclidean space. However, there are a lot of applications where data are generated from the non-Euclidean domain such as graphs. GraphCNN is recently widely studied to generalize CNN to graph structured data \cite{GCN}. Fig. \ref{GCN} shows an structure illustration of a GraphCNN model. The “graph convolution” operation applies the convolutional transformation to the neighbors of each node, followed by pooling operation. By stacking multiple graph convolution layers, the latent embedding of each node can contain more information from neighbors which are multi-hops away. After the generation of the latent embedding of the nodes in the graph, one can either easily feed the latent embeddings to feed-forward networks to achieve node classification of regression goals, or aggregate all the node embeddings to represent the whole graph and then perform graph classification and regression. Due to its powerful ability in capturing the node correlations as well as the node features, it is now widely used in mining graph structured ST data such as network-scale traffic flow data and brain network data.

\begin{figure}[t]
\centering  
\subfigure[RNN]{
\label{rnn}
\includegraphics[height=2cm, angle=0]{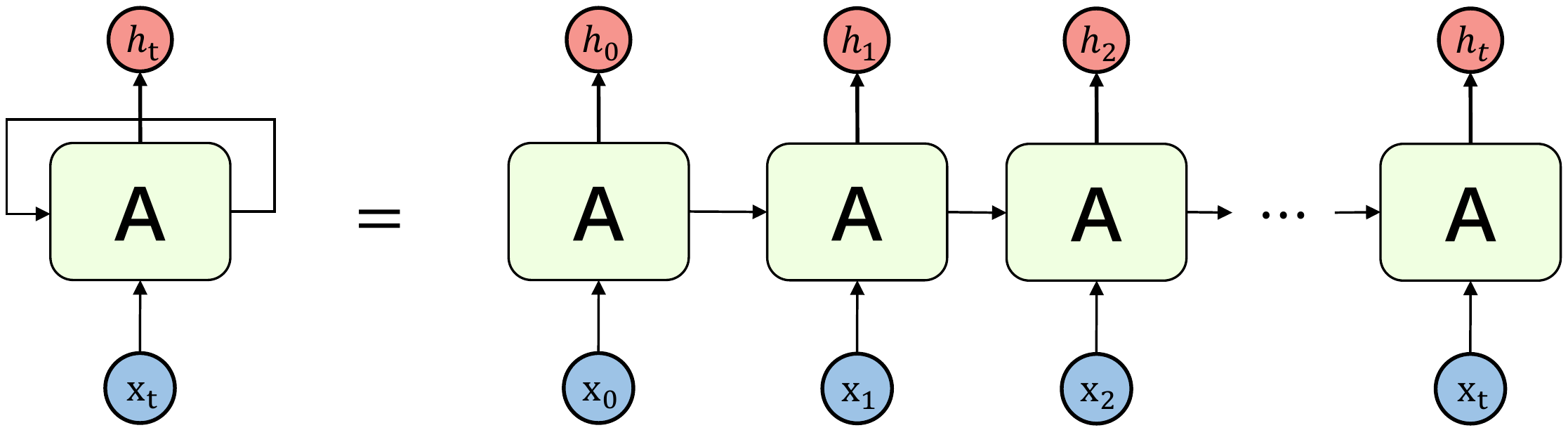}}
\subfigure[LSTM]{
\label{LSTM}
\includegraphics[height=1.6cm, angle=0]{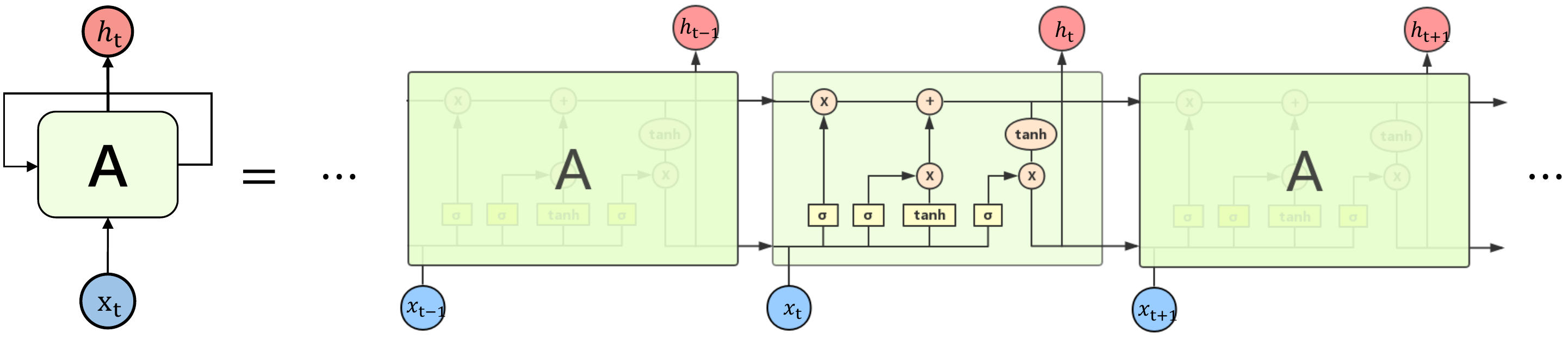}}
\caption{Structure of the RNN and LSTM models}
\end{figure}
\textbf{RNN and LSTM.} A recurrent neural network (RNN) is a class of artificial neural network where connections between nodes form a directed graph along a sequence. RNN is designed to recognize the sequential characteristics and use patterns to predict the next likely scenario. They are widely used in  the applications of speech recognition and natural language processing. Fig. \ref{rnn} shows the general structure of a RNN model, where ${X}_t$ is the input data, ${A}$ is the parameters of the network and $h_t$ is the learned hidden state. One can see the output (hidden state) of the previous time step $t-1$ is input into the neural of the next time step $t$. Thus the historical information can be stored and passed to the future. 

A major issue of standard RNN is that it only has short-term memory due to the issue of vanishing gradients. Long Short-Term Memory (LSTM) network is an extension for recurrent neural networks, which is capable of learning long-term dependencies of the input data. LSTM enables RNN to remember their inputs over a long period of time due to the special memory unit as shown in the middle part of Fig. \ref{LSTM}. An LSTM unit is composed of three gates: input, forget and output gate. These gates determine whether or not to let new input in (input gate), delete the information because it is not important (forget gate) or to let it impact the output at the current time step (output gate). Both RNN and LSTM are widely used to deal with sequence and time serious data for learning the temporal dependency of the ST data.

\begin{figure}[!t]
\begin{center}
\includegraphics[height=3.5cm]{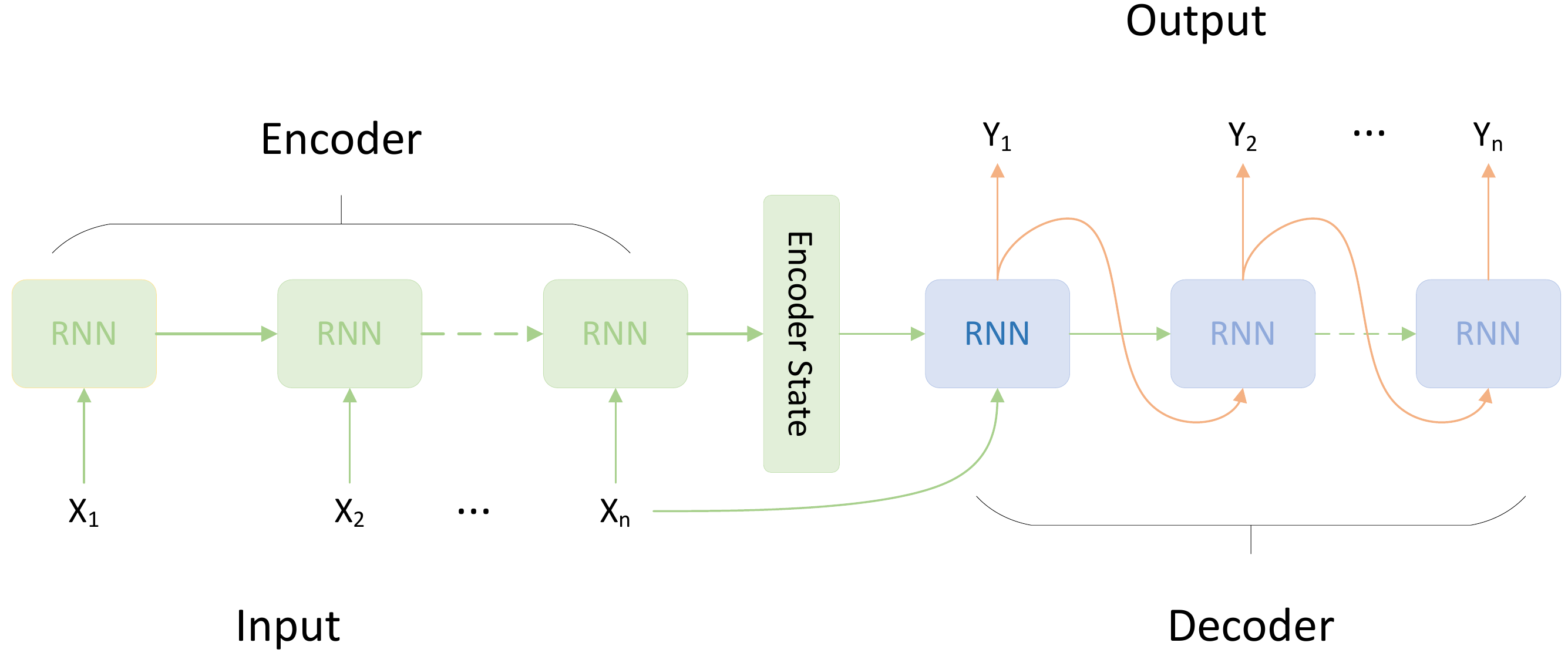}
\caption{Structure of Seq2Seq model.}
\label{seq}
\end{center}
\end{figure}
\textbf{Seq2Seq.} A sequence to sequence (Seq2Seq) model aims to map a fixed length input with a fixed length output where the length of the input and output may differ \cite{seq2seq}. It is widely used to various NLP tasks such as machine translation, speech recognition and online chatbot. Although it is initially proposed to address NLP tasks, Seq2Seq is general framework and can be used to any sequence-based problem. As shown in Fig. \ref{seq}, a Seq2Seq model generally consists of 3 parts: encoder, intermediate (encoder) vector and decoder. Due to the powerful ability in capturing the dependencies among the sequence data, Seq2Seq model is widely used in ST prediction tasks where the ST data present high temporal correlations such as urban crowd flow data and traffic data.

\begin{figure}[t]
\begin{center}
\includegraphics[height=4.5cm, angle=0]{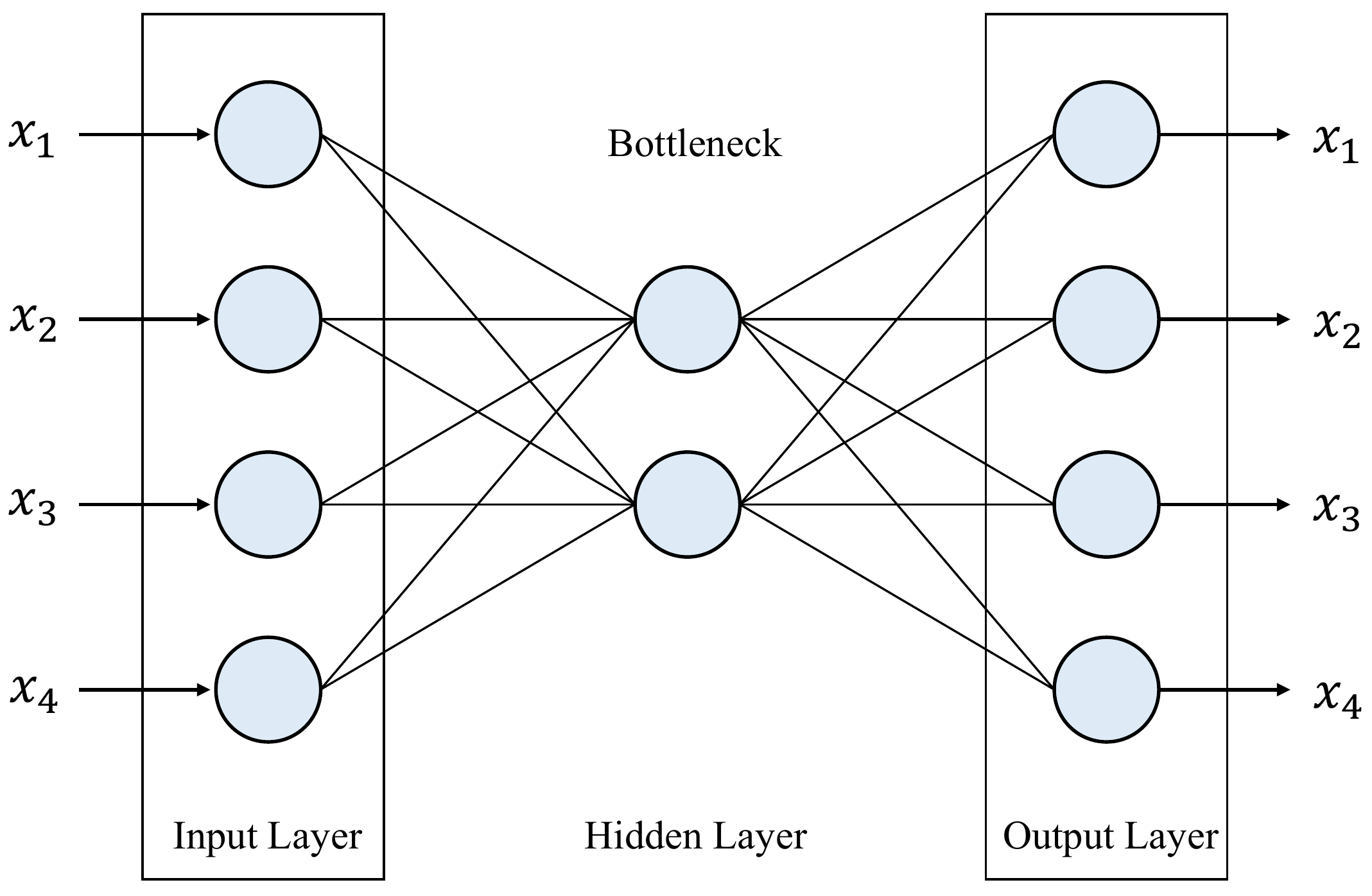}
\caption{Structure of the one-layer AE model.}
\label{ae}
\end{center}
\end{figure}

\textbf{Autoencoder (AE) and Stacked AE.} An autoencoder is a type of artificial neural network that aims to learn efficient data codings in an unsupervised manner \cite{RBM}. As shown in Fig. \ref{ae}, it features an encoder function to create a hidden layer (or multiple layers) which contains a code to describe the input. There is then a decoder which creates a reconstruction of the input from the hidden layer. An autoencoder creates a compressed representation of the data in the hidden layer or bottleneck layer by learning correlations in the data, which can be considered as a way for dimensionality reduction. As an effective unsupervised feature representation learning technique, AE facilitates various down stream data mining and machine learning tasks such as classification and clustering. A stacked autoencoder (SAE) is a neural network consisting of multiple layers of sparse autoencoders in which the outputs of each layer is wired to the inputs of the successive layer \cite{SAE}.

\section{Framework}
In this section, we will introduce how to use deep learning models for addressing STDM problems in general. First, we will give a framework that describes the pipeline which contains ST data instance construction, ST data representation, deep learning model section \& design, and finally addressing the problem. Next we will introduce these major steps in detail.
\begin{figure}[!t]
\begin{center}
\includegraphics[height=5.5cm]{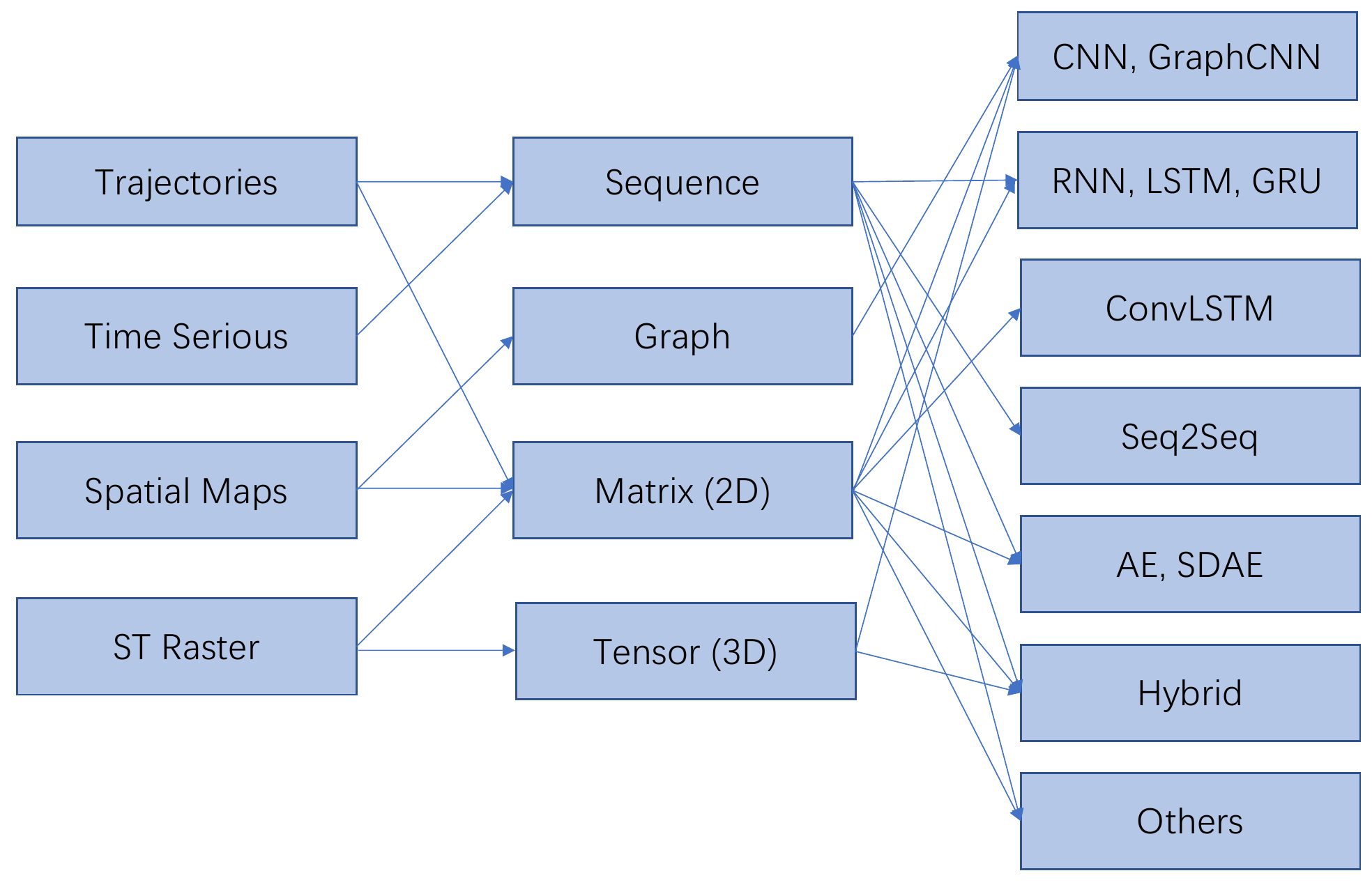}
\caption{Data representation for different DL models}
\label{Rep}
\end{center}
\end{figure}

\begin{figure*}[htbp]
\begin{center}
\includegraphics[height=3.8cm]{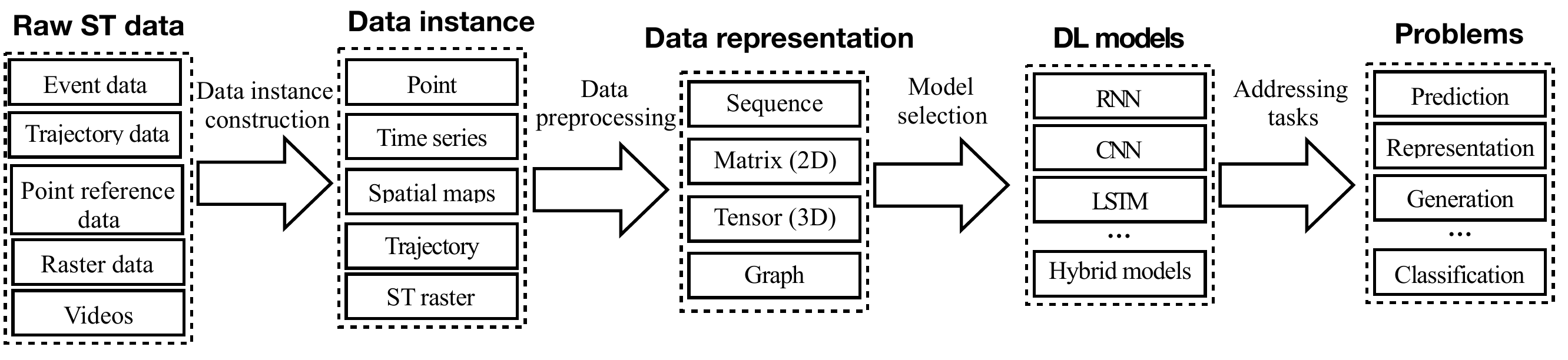}
\caption{A general pipeline for using DL models for ST data mining}
\label{framework}
\end{center}
\end{figure*}

A general pipeline for using deep learning models for ST data mining is shown in Fig. \ref{framework}. Given the raw ST data collected from various location sensors, including the event data, trajectory data, point reference data and raster data, data instances are first constructed for data storage. As we discussed before, the ST data instances can be point, time series, spatial maps, trajectory and ST raster. To apply deep learning models for various mining tasks, the ST data instances need to be further represented as a particular data format to fit the deep learning models. The ST data instances can be represented as sequence data, 2D matrix, 3D tensors and graphs. Then for different data representations, different deep learning models are suitable to process them. RNN and LSTM models are good at handling sequence data with short-term or long-term temporal correlation, while CNN models are effective to capture the spatial correlation in the image like matrices. The hybrid model that combines RNN and CNN can capture both the spatial and temporal correlations of a tensor representation of the ST raster data. Finally, the selected deep learning models are used to address various STDM tasks such as prediction, classification, representation learning, etc. 

\subsection{ST Data Preprocessing}
ST data preprocessing aims to represent ST data instances as a proper data representation format that the deep learning model can handle. Usually the input data format of a deep learning model can be a vector, a matrix or a tensor depending on different models. Fig. \ref{Rep} shows the ST data instances and their corresponding data representations. One can see that usually one type of ST data instance corresponds to one typical data representations. Trajectory and time series data can be naturally represented as sequence data. Spatial map data can be represented as a 2D matrix. ST raster can be represented as a 2D matrix or 3D tensor. 

However, it is not always the case. For example, trajectory data sometimes are represented as a matrix, and then CNN model is applied to better capture the spatial features \cite{dabiri2018inferring,karatzoglou2018convolutional,lv2018t,niu2014deepsense,wang2017ridesourcing}. The ST field where the trajectories are measured such as a city is first partitioned into grid cell regions. Then the ST field can be modeled as a matrix with each cell region representing an entry. If a trajectory paths over the cell region, the corresponding entry value is set to 1; otherwise it is set to 0. In this way, a trajectory data can be represented as a matrix and thus CNN can be applied. Sometimes a spatial map is represented as a graph. For example, the sensors deployed in the express ways are usually modeled as a graph where the nodes are the sensors and the edges denote the road segments between two neighbor sensors. In such a case, GraphCNN models are usually utilized to process the sensor graph data and predict the future traffic (volume, speed, etc.) for all the nodes \cite{cui2018high,li2018diffusion}. ST raster data can be both represented as 2D matrices or 3D tensors, depending on the data types and applications. For example, a series of fMRI brain image data can be represented as a tensor and input into a 3D-CNN model for diseases classification \cite{korolev2017residual,nie20163d}, and it can be also represented as a matrix by extracting the time series correlations between pair-wise regions of the brain for brain activity analysis \cite{guo2017diagnosing,meszlenyi2017resting}.

\subsection{Deep Learning Model Selection \& Design}
With the data representations of the ST data instances, the next step is to feed them into the selected or designed deep learning models for different STDM tasks. As shown in the right part of Fig. \ref{Rep}, there are different deep learning model options for each type of data representation. Sequence data can be used as the input of the models including RNN, LSTM, GRU, Seq2Seq, AE, hybrid models and others. RNN, LSTM and GRU are all recurrent neural networks that are suitable to predict the sequence data. Sequence data can also be processed by Seq2Seq model. For example, in multi-step traffic prediction, a Seq2Seq model which consists a set of LSTM units in the encoder layer and a set of LSTM units in the decoder layer is usually applied to predict the traffic speed or volume in the next several time slots simultaneously \cite{liao2018dest,liao2018deep}. As a feature learning model, AE or SAE can be used to various data representations to learn a low-dimensional feature coding. Sequence data can also be encoded as a low-dimensional feature with AE or SAE. GraphCNN is particularly designed to process the graph data to capture the spatial correlations among the neighbor nodes. If the input is a single matrix, usually CNN model is applied, and if the input is a sequence of matrices, RNN models, ConvLSTM and hybrid models can be applied depending on the problems under study. If the goal is only for feature learning, AE and SAE models can be applied. For tensor data, usually it is handled by a 3D-CNN or the combination of 3D-CNN with RNN models.  

Table \ref{modelfordata} summarizes the works using deep learning models to handle different types of ST data. As shown in the table, CNN, RNN and their variants (e.g. GraphCNN and ConvLSTM) are two most widely used deep learning models for STDM. CNN model is mostly used to process the spatial maps and ST raster. Some works also used CNN to handle trajectory data, but currently there is no work using CNN for time series data learning. GraphCNN model is specially designed to handle graph data, which can be categorized into spatial maps. RNN models including LSTM and GRU can be broadly applied in dealing with trajectories, time series, and the sequences of spatial maps. ConvLSTM can be considered as a hybrid model which combines RNN and CNN, and are usually used to handle spatial maps. AE and SDAE are mostly used to learn features from time series, trajectories and spatial maps. Seq2Seq model is generally designed for sequential data, and thus only used to handle time series and trajectories. The hybrid models are also common for STDM. For example, CNN and RNN can be stacked to learn the spatial features first, and then capture the temporal correlations among the historical ST data. Hybrid models can be designed to fit all the four types of data representations. Other models such as network embedding \cite{yang2017neural}, multi-layer perceptron (MLP) \cite{huang2018deepcrime,zhang2018short}, generative adversarial nets (GAN) \cite{gupta2018social,lin2018pattern}, Residual Nets \cite{korolev2017residual,liao2018dest}, deep reinforcement learning \cite{WWW19MoD}, etc. are also used in recent works.

\subsection{Addressing STDM Problems}
Finally, the selected or designed deep learning models are used to address various STDM tasks such as classification, predictive learning, representation learning and anomaly detection. Note that usually how to select or design a deep learning model depends on the particular data mining task and the input data. However, to show the pipeline of the framework we first show the deep learning model and then the data mining tasks. In next section, we will categorize different STDM problems and review the works based on the problems and ST data types in detail.

\begin{table*}[htp]
\caption{Different DL models for processing four types of ST data.}
\begin{center}
\begin{tabular}{|p{3cm}|p{3cm}|p{3cm}|p{3cm}|p{3cm}|p{3cm}|}
\hline
 & Trajectories & Time Series & Spatial Maps (Image-like data \& Graphs) & ST Raster\\
\hline
CNN &\cite{dabiri2018inferring,karatzoglou2018convolutional,lv2018t,niu2014deepsense,wang2017ridesourcing} & &\cite{chattopadhyay2018test,wang2016action,zhu2018deep,wang2015action,liu2016application,duan2017deep,tao2016deep,wang2017detecting,zhang2016dnn,lee2018forecasting,ke2018hexagon,chen2018pcnn,kim2017resolution,zhu2018wind,meszlenyi2017resting,horikawa2017generic,kawahara2017brainnetcnn}&\cite{zhang2017application,chen2018exploiting,racah2017extremeweather,tran2015learning,ma2017learning,kira2016leveraging,shen2018stepdeep,wang2016traffic,nie20163d,sarraf2016deep,sarraf2016deep,kleesiek2016deep,korolev2017residual}\\
\hline
GraphCNN & & &\cite{li2018diffusion,wang2018efficient,lin2018exploiting,martin2018graph,wang2018graph,cui2018high,lin2018predicting,yuspatio,geng2019spatiotemporal,chai2018bike,li2018diffusion,wang2018efficient} &\\
\hline
RNN(LSTM,GRU) &\cite{gao2017identifying,kong2018hst,yang2018recurrent,liu2016predicting,liao2018proceedings,xu2018collision,endo2017predicting,wu2017modeling,jiang2018deep,feng2018deepmove,song2016deeptransport,zhang2018deeptravel,liang2018geoman,li2018next,zhao2018prediction,fan2018online,yao2017serm,yang2018spatio,gao2018trajectory,jiang2017trajectorynet,zhao2018go} & \cite{rodrigues2019combining,Dixon2017Deep,yu2017deep,liao2018deep,cui2016deep,liao2018dest,rong2018parking,cheng2018ensemble,zaytar2016sequence,liu2017short,liu2018smart1,chen2017using,dvornek2017identifying} & \cite{ren2018deep,ma2018forecasting,wang2017predrnn,akbari2018short,alahi2016social,fernando2018soft+,jain2016structural,xu2018station}&\cite{cui2016deep}\\
\hline
ConvLSTM & & &\cite{ai2018deep,liu2018attentive,xingjian2015convolutional,zhou2018predicting,DBLP,kim2017deeprain,yuan2018hetero,ke2017short,wang2018will} &\\
\hline
AE/SDAE &\cite{nguyen2012extracting,zhou2018trajectory,chen2018trip2vec} &\cite{hossain2015forecasting,yang2017optimized,lv2015traffic} & \cite{duan2014deep,chen2016learning,zhao2018temporal,guo2017diagnosing,heinsfeld2018identification,zhang2016identifying} &\\
\hline
RBM/DBN &\cite{niu2014deepsense} & \cite{soua2016big}& &\cite{taylor2010convolutional,huang2016latent,jin2014classification}\\
\hline
Seq2Seq &\cite{li2018deep,yao2018learning,chow2018representation,yao2017trajectory} & \cite{liao2018deep,liao2018dest} & &\\
\hline
Hybrid &\cite{yang2017neural,varshneya2017human,ma2015large} & \cite{liu2018smart,huang2018modeling}&\cite{zhang2018deep,du2018hybrid,cheng2018neural,bao2019spatiotemporal,yao2018deep,zhang2017fcn,li2018origin,ma2018parallel,shi2017sequential,gupta2018social,yu2017spatiotemporal} &\cite{lv2018lc,roesch2017visualization}\\
\hline
Others & \cite{endo2016classifying,chang2018content,giffard2018fused,zheng2016generating,ding2018geographical,zhao2017geo,yang2018unsupervised} &\cite{rasp2018neural,lin2018pattern} & \cite{shi2017deep,wang2017deep1,zonoozi2018periodic,costilla2018deep,zhang2017deep,wang2017deepsd,kurth2018exascale,gao2019incomplete,zhang2018predicting,zhang2018short,shi2018multimodal}&\cite{qiu2017learning,jang2017task,kim2016deep}\\
\hline
\end{tabular}
\end{center}
\label{modelfordata}
\end{table*}%

\section{Deep Learning Models for Addressing Different STDM Problems}
\begin{figure}[!h]
\begin{center}
\includegraphics[height=5.5cm]{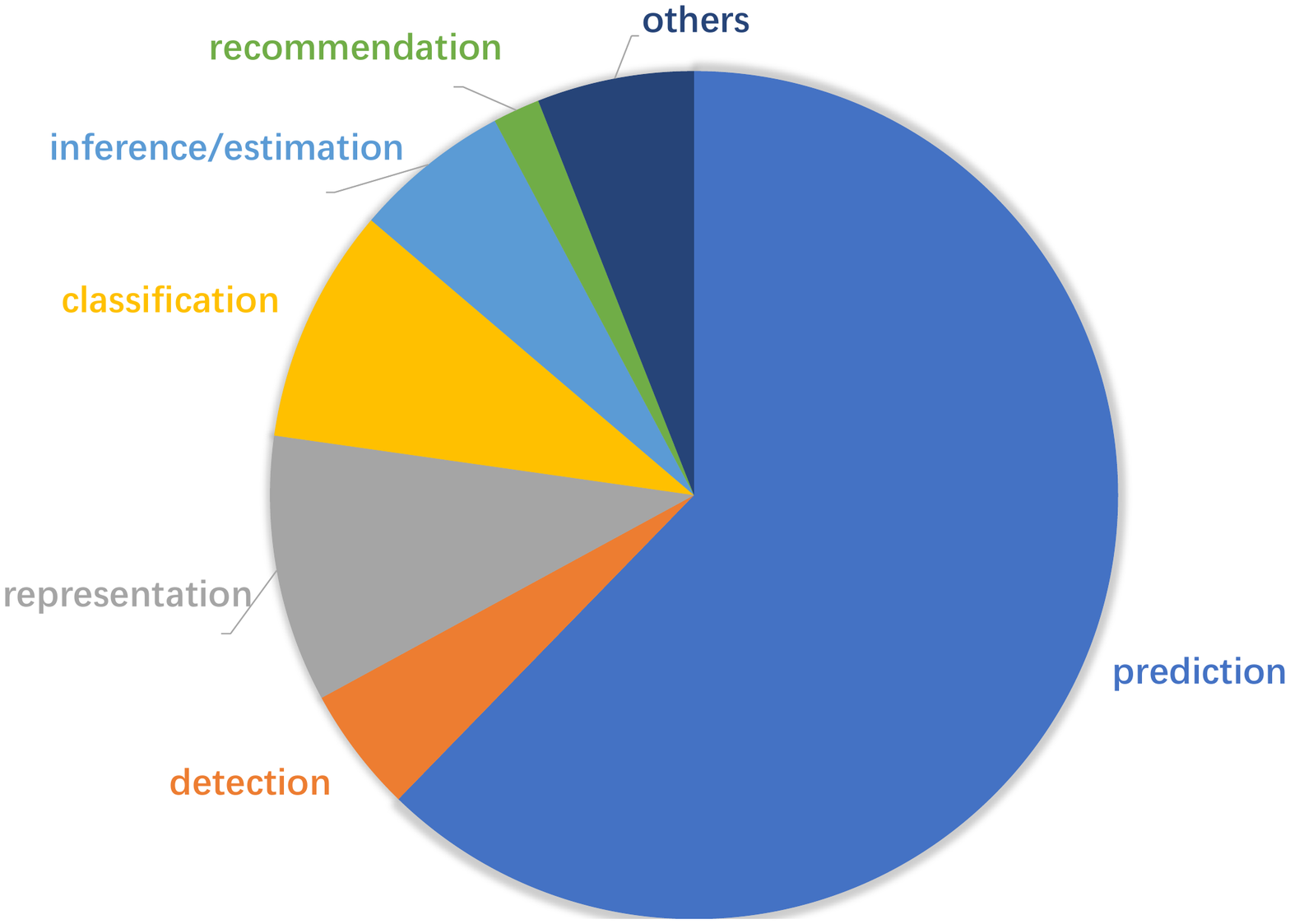}
\caption{Distributions of the STDM problems addressed by deep learning models}
\label{Pro}
\end{center}
\end{figure}

In this section, we will categorize the STDM problems, and introduce the corresponding deep learning models proposed to address them. Fig. \ref{Pro} shows the distribution of various STDM problems addressed by deep learning models, including prediction, representation learning, detection, classification, inference/estimation, recommendation and others. One can see the largest category of the studied STDM problems is prediction. More than 70\% related papers focus on studying the ST data prediction problem. This is mainly because an accurate prediction largely relies on high quality features, while deep learning models are especially powerful in feature learning. The second largest problem category is representation learning, which aims to learning feature representations for various ST data in an unsupervised or semi-supervised way. Deep learning models are also used in other STDM tasks including classification, detection, inference/estimation, recommendation, etc. Next we will introduce the major STDM problems in detail and summarize the corresponding deep learning based solutions.

\subsection{Predictive Learning}
The basic objective of predictive learning is to predict the future observations of the ST data based on its historical data. For different applications, both the input and output variables can belong to different types of ST data instances, resulting in a variety of predictive learning problem formulations. In the following, we will introduce the predictive problems based on the types of ST data instance as the model input.

\textbf{Points.} Points are usually merged in temporal or spatial domains to form time series or spatial maps such as crimes \cite{duan2017deep,huang2018deepcrime,wang2017deep1,WWW19Mist}, traffic accidents \cite{yuan2018hetero} and social events \cite{gao2019incomplete}, so that deep learning models can be applied. \cite{wang2017deep1} adapted ST-ResNet model to predict crime distribution over the Los Angeles area. Their models contains two staged. First, they transformed the raw crime point data as image-like crime heat maps by merging all the crime events happened in the same time slot and region of the city. Then, they adapted hierarchical structures of residual convolutional units to train a crime prediction model with the crime heat maps as input. Similarly, \cite{huang2018deepcrime} proposed to use GRU model to predict the crime of a city. \cite{yuan2018hetero} studied the traffic accident prediction problem using the Convolutional Long Short-Term Memory (ConvLSTM) neural network model. They also first merged the point data of traffic accidents and modeled the traffic accident count in a spatio-temporal field as a 3-D tensor.  Each entry $(i,j,t)$ of the tensor represents the traffic accident count at the grid cell $(i,j)$ in time slot $t$. The historical traffic accident tensors are input into CovnLSTM for prediction. \cite{gao2019incomplete} proposed a spatial incomplete multi-task deep learning framework to effectively forecast the subtypes of future events happened at different locations.

\textbf{Time series.} In road-level traffic prediction, the traffic flow data on a road or freeway can be modeled as a time series. Recently, many works tried various deep learning models for road-level traffic prediction \cite{lv2015traffic,soua2016big,zhao2018temporal}. \cite{lv2015traffic} for the first time utilized stacked autoencoder to learn features from the traffic flow time series data for road-segment level traffic flow prediction. \cite{soua2016big} considered the traffic flow data at a freeway as time series and proposed to use Deep Belief Networks (DBNs) to predict the future traffic flow based on the previous traffic flow observations. \cite{rodrigues2019combining} studied the problem of taxi demand forecasting, and modeled the taxi demand at a particular area as a time series. A deep learning model with fully-connected layers is proposed to learn features from the historical time series of taxi demand, and then the features are integrated with other context features such as weathers and social media texts to predict the future demand. 

RNN and LSTM are widely used for time series ST data prediction. \cite{liao2018deep} integrated LSTM and sequence to sequence model to predict the traffic speed of a road segment. Besides the traffic speed information, their model also considered other external features including the geographical structure of roads, public social events such as national celebrations, and online crowd queries for travel information. The weather variables such as wind speed are also typically modeled as time series and then RNN/LSTM models are applied for future weather forecasting \cite{chen2017using,cheng2018ensemble,hossain2015forecasting,liu2018smart1,rasp2018neural,zaytar2016sequence}. For example, \cite{cheng2018ensemble} proposed an ensemble model for probabilistic wind speed forecasting. The model integrated traditional wind speed prediction models including wavelet threshold denoising (WTD) and adaptive neuro fuzzy inference system (ANFIS) with recurrent neural network (RNN). In the area of fMRI data analysis, fMRI time series data are usually used to study the functional brain network and diagnose disease. \cite{dvornek2017identifying} proposed to use LSTM model for classification of individuals with autism spectrum disorders (ASD) and typical controls directly from the resting-state fMRI time-series. \cite{huang2018modeling}  developed a deep convolutional auto-encoder model named DCAE for learning mid-level and high-level features from complex, large-scale tfMRI time series in an unsupervised manner. The time series data usually do not contain the spatial information, and thus the spatial correlations among the data are not explicitly considered in deep learning based prediction models.

\textbf{Spatial maps.} 
The spatial maps can be usually represented as image-like matrices, and thus are suitable to be processed with CNN models for predictive learning \cite{ke2018hexagon,lee2018forecasting,zhang2016dnn,zhu2018wind}. \cite{zhang2016dnn} proposed a CNN based prediction model to capture the spatial features in urban crow flow prediction. A real-time crowd flow forecasting system called UrbanFlow is built, and the crowd flow spatial maps are as its input. For forecasting the supply-demand in ride-sourcing services, \cite{ke2018hexagon} proposed the hexagon-based convolutional neural networks (H-CNN), where the input and output are both numerous local hexagon maps. In contrast to the previous studies that partitioned a city area into numerous square lattices, they proposed to partition the city area into various regular hexagon lattices because hexagonal segmentation has an unambiguous neighborhood definition, smaller edge-to-area ratio, and isotropy. Wind speed data of one monitoring site can be modeled as time series, while the data of multiple sites can be represented as spatial maps. CNN models can be also applied to predict wind speed of multiple sites simultaneously \cite{zhu2018wind}. 

Given a sequence of spatial maps, to capture both the temporal and spatial correlations many works tried to combine CNN with RNN for the prediction. \cite{xingjian2015convolutional} proposed a convolutional LSTM (ConvLSTM) and used it to build an end-to-end trainable model for the precipitation nowcasting problem. This work combined the convolutional structure in CNN and the LSTM unites to predict the spatio-temporal sequences under a sequence-to-sequence learning framework. ConvLSTM is a sequence-to-sequence prediction model, whose each layer is a ConvLSTM unit that has convolutional structures in both the input-to-state and state-to-state transitions. The input and output of the model are both spatial map matrices. Following this work, many works tried to apply ConvLSTM to other spatial map prediction tasks of different domains \cite{ai2018deep,bao2019spatiotemporal,TITS19,ke2017short,kim2017deeprain,liu2018attentive,DBLP,zhou2018predicting}. \cite{DBLP} proposed a novel cross-city transfer learning method for deep spatio-temporal prediction, called \emph{RegionTrans}. \emph{RegionTrans} contained multiple ConvLSTM layers to catch the spatio-temporal patterns hidden in the data. \cite{kim2017deeprain} applied ConvLSTM network to predict precipitation by using multi-channel radar data. \cite{zhou2018predicting} proposed an end-to-end deep neural network for predicting the passenger pickup/dropoff demands in mobility-on-demand (MOD) service. A encoder-decoder framework based on convolutional and ConvLSTM units was employed to identify complex features that capture spatio-temporal influences and pickup-dropoff interactions on citywide passenger demands. The passenger demands in the cell regions of a city was modeled as a spatial map and represented as a matrix. Similarly, \cite{ai2018deep} proposed a FCL-Net model which fused ConvLSTM layers, standard LSTM layers and convolutional layers for forecasting of passenger demand under on-demand ride services. \cite{liu2018attentive} proposed a unified neural network module called Attentive Crowd Flow Machine (ACFM). ACFM is able to infer the evolution of the crowd flow by learning dynamic representations of temporally-varying data with an attention mechanism. ACFM is composed of two progressive ConvLSTM units connected with a convolutional layer for spatial weight prediction.

Some other models can be also used for predicting spatial maps, such as GraphCNN \cite{chai2018bike,cui2018high,lin2018predicting,wang2018graph}, ResNet \cite{wang2017deepsd,zhang2017deep,zhang2018predicting}, and hybrid methods \cite{gupta2018social,ma2018parallel,yu2017deep}. Note in this paper we consider that spatial maps contain both image-like data and graph data. Although graphs are also represented as matrices, they require totally different technique such as GraphCNN or GraphRNN. In road network-scale traffic prediction, the transportation network can be naturally modeled as a graph, and then GraphCNN or GraphRNN is applied. \cite{li2018diffusion} proposed to model the traffic flow on a transportation network as a diffusion process on a directed graph and introduced Diffusion Convolutional Recurrent Neural Network (DCRNN) for traffic forecasting. It incorporated both spatial and temporal dependency in the traffic flow of the entire road network. Specifically, DCRNN captures the spatial dependency using bidirectional random walks on the graph, and the temporal dependency using the encoder-decoder architecture with scheduled sampling. \cite{wang2018efficient} proposed a new topological framework called Linkage Network to model the road networks and presented the propagation patterns of traffic flow. Based on the Linkage Network model, a novel online predictor, named Graph Recurrent Neural Network (GRNN), is designed to learn the propagation patterns in the graph. It simultaneously predicts traffic flow for all road segments based on the information gathered from the whole graph. \cite{wang2018graph} introduced an ST weighted graph (STWG) to represent the sparse spatio-temporal data. Then to perform micro-scale forecasting of the ST data, they built a scalable graph structured RNN (GSRNN) on the STWG.

\textbf{Trajectories.} Currently, two types of deep learning models, RNN and CNN are used for trajectory prediction depending on the data representations of the trajectories. First, trajectories can be represented as the sequence of locations as shown in Fig. \ref{Rep}. In such a case, RNN and LSTM models can be applied \cite{feng2018deepmove,jiang2018deep,kong2018hst,liang2018geoman,song2016deeptransport,xu2018collision,yang2018recurrent}. \cite{xu2018collision} proposed Collision-Free LSTM, which extended the classical LSTM by adding Repulsion pooling layer to share hidden-states of neighboring pedestrians for human trajectory prediction. Collision-Free LSTM can generate the future sequence based on pedestrian past positions. \cite{jiang2018deep} studied the urban human mobility prediction problem, which given a few steps of observed mobility from one person, tries to predict where he/her will go next in a city. They proposed a deep-sequence learning model with RNN to effectively predict urban human mobility. \cite{song2016deeptransport} proposed a model named DeepTransport to predict the transportation mode such as walk, taking train, taking bus, etc, from a set of individual people’s GPS trajectories. Four LSTM layers are used to constructed DeepTransport to predict a user's transportation mode in the future.

Trajectories can be also represented as a matrix. In such a case, CNN models can be applied to better capture the spatial correlations \cite{karatzoglou2018convolutional,lv2018t,varshneya2017human}. \cite{karatzoglou2018convolutional} proposed a CNN-based approach for representing semantic trajectories and predicting future locations. In a semantic trajectory, each visited location is associated with a semantic meaning such as \emph{home}, \emph{work}, \emph{shoppint}, etc. They modeled the semantic trajectories as a matrix whose two dimensions are semantic meanings and trajectory ID. The matrix is input into a CNN with multiple convolutional layers to learn the latent features for next visited semantic location prediction. \cite{lv2018t} modeled trajectories as two-dimensional images, where each pixel of the image represented whether the corresponding location was visited in the trajectory. Then multi-layer convolutional neural networks were adopted to combine multi-scale trajectory patterns for destination prediction of taxi trajectories. Modeling trajectories as image-like matrix is also utilized in other tasks such anomaly detection and inference \cite{martin2018graph,wang2017ridesourcing}, which will be introduced in detail later.

\textbf{ST raster.} As we discussed before, ST raster data can be represented as matrices whose two dimensions are location and time, or tensors whose three dimensions are cell region ID, cell region ID, and time. Usually for ST raster data prediction, 2D-CNN (matrices) and 3D-CNN (tensors) are applied, and sometimes they are also combined with RNN. \cite{zhang2017application} proposed a multi-channel 3D-cube successive convolution network named 3D-SCN to nowcast storm initiation, growth, and advection from the 3D radar data. \cite{polson2017deep} modeled the traffic speed data at multiple locations of a road in successive time slots as a ST raster matrix, and then input it into a deep neural network for traffic flows prediction. \cite{ma2017learning} explored the similar idea as \cite{polson2017deep} for traffic prediction on a large transportation network. \cite{chen2018exploiting} proposed a 3D Convolutional neural networks for citywide vehicle flow prediction. Instead of predicting traffic on a road, they tried to predict vehicle flows in each cell region of a city. So they modeled the citywide vehicle flow data in successive time slots as ST rasters and input them into the proposed 3D-CNN model. Similarly, \cite{shen2018stepdeep} modeled the mobility events of passengers in a city in different time slots as a 3D tensor, and then used 3D-CNN model to predict the supply and demand of the passengers for transportation. Note that the major difference between ST raster and spatial maps is that ST raster is the merged ST field measurements of multiple time slots, while spatial map is the ST field measurement in only one time slot. Thus the same type of ST data sometimes can be represented as both spatial maps and ST raster depending on the real application scenarios and the purposes of data analysis. 

\subsection{Representation Learning}
Representation learning aims to learn the abstract and useful representations of the input data to facilitate downstream data mining or machine learning tasks, and the representations are formed by composition of multiple linear or non-linear transformations of the input data. Most existing works on representation learning for ST data focused on studying the data types of trajectories and spatial maps.

\textbf{Trajectories.} Trajectories are ubiquitous in location-based social networks (LBSNs) and various mobility services, and RNN and CNN models are both widely used to learn the trajectory representations. \cite{li2018deep} proposed a seq2seq-based model to learn trajectory representations, for the fundamental research problem of trajectory similarity computation. The trajectory similarity based on the learned representations is robust to non-uniform, low sampling rates and noisy sample points. Simiarly, \cite{yao2018learning,yao2017trajectory} proposed to transform a trajectory into a feature sequence to describe object movements, and then employed a sequence‐to‐sequence auto‐encoder to learn fixed‐length deep representations for clustering. Location-based social network (LBSN) data usually contain two important aspects, i.e., the mobile trajectory data and the social network of users. To model the two aspects and mine their correlations, \cite{yang2017neural} proposed a neural network model to jointly learn the social network representation and the users' mobility trajectory representation. RNN and GRU models are used to capture the sequential relatedness in mobile trajectories at the short or long term levels. \cite{chang2018content} proposed a content-aware POI embedding model named CAPE for POI recommendation. In CAPE, the embedding vectors of POIs in a user's check-in sequence are trained to be close to each other. \cite{ding2018geographical} proposed a geographical convolutional neural tensor network named GeoCNTN to learn the embeddings of the locations in LBSNs. \cite{gao2018trajectory} proposed to use RNN and Autoencoder to learn the user check-in embedding and trajectory embedding, and used the embeddings for user social circle inference in LBSNs.

\textbf{Spatial maps.} There are several works that study how to learn representations of the spatial maps. \cite{costilla2018deep} proposed a convolutional neural network architecture for learning spatio-temporal features from raw spatial maps of the sensor data. \cite{wang2018learning} formulated the problem of learning urban community structures as a spatial representation learning task. A collective embedding learning framework was presented to learn urban community structures by unifying both static POIs data and dynamic human mobility graph spatial map data. \cite{zhang2016identifying} studied how to learn nonlinear representations of brain connectivity patterns from neuroimage data to inform an understanding of neurological and neuropsychiatric disorders. A deep learning architecture named Multi-side-View guided AutoEncoder (MVAE) is proposed to learn the representations of the input brain connectome data derived from fMRI and DTI images.


\subsection{Classification}
The classification task is mostly studied in analyzing fMRI data. Recently, brain imaging technology has become a hot topic within the field of neuroscience, including functional Magnetic Resonance Imaging (fMRI), electroencephalography (EEG), and Magnetoencephalography (MEG) \cite{plis2014deep}. Particularly, fMRI combined with deep learning methods, has been widely used in the study of neuroscience for various classification tasks such as disease classification, brain function network classification and brain activation classification when watching words or images \cite{wen2018deep}. Various types of ST data can be extracted from the raw fMRI data depending on different classification tasks. \cite{dvornek2017identifying} proposed the use of recurrent neural networks with long short-term memory (LSTMs) for classification of individuals with autism spectrum disorders (ASD) and typical controls directly from the resting-state fMRI time-series data generated from different brain regions. \cite{guo2017diagnosing,heinsfeld2018identification,horikawa2017generic,kim2016deep,meszlenyi2017resting,shi2018multimodal} modeled the fMRI data as spatial maps, and then used them as the input of the classification models. Instead of using each individual resting-state fMRI time-series data directly, \cite{guo2017diagnosing} and \cite{heinsfeld2018identification} calculated the whole-brain functional connectivity matrix based on the Pearson correlation coefficient between each pair of resting-state fMRI time-series data. Then the correlation matrix can be considered as a spatial map, and is input to a DNN model for ASD classification. \cite{meszlenyi2017resting} proposed a more general convolutional neural network architecture for functional connectome classification called connectome-convolutional neural network (CCNN). CCNN is able to combine information from diverse functional connectivity metrics, and thus can be easily adapted to a wide range of connectome based classification or regression tasks, by varying which connectivity descriptor combinations are used to train the network. 

Some works also directly use the 3D structural MRI brain scanned images as the ST raster data, and then 3D-CNN model is usually applied to learn features from the ST raster for classification \cite{jang2017task,jin2014classification,korolev2017residual,nie20163d,sarraf2016deep,zhao2018automatic}. \cite{korolev2017residual} proposed two 3D convolutional network architectures for brain MRI classification, which are the modifications of a plain and residual convolutional neural networks. Their models can be applied to 3D MRI images without intermediate handcrafted feature extraction. \cite{zhao2018automatic} also designed a deep 3D-CNN framework for automatic, effective, and accurate classification and recognition of large number of functional brain networks reconstructed by sparse 3D representation of whole-brain fMRI signals.
 
 \subsection{Estimation and Inference}
 Current works on ST data estimation and inference mainly focus on the data types of spatial maps and trajectories.
 
 \textbf{Spatial maps.} While monitoring stations have been established to collect pollutant statistics, the number of stations is very limited due to the high cost. Thus, inferring fine-grained urban air quality information is becoming an essential issue for both government and people. \cite{cheng2018neural} studied the problem of air quality inference for any location based on the air pollutant of some monitoring stations. They proposed a deep neural network model named ADAIN for modeling the heterogeneous data and learning the complex feature interactions. In general, ADAIN combines two kinds of neural networks: i.e., feedforward neural networks to model static data and recurrent neural networks to model sequential data, followed by hidden layers to capture feature interactions. \cite{tao2016deep} investigated the application of deep neural networks to precipitation estimation from remotely sensed information. A stacked denoising auto-encoder is used to automatically extract features from the infrared cloud images and estimate the amount of precipitation. Estimating the duration of a potential trip given the origin location, destination location as well as the departure time is a crucial task in intelligent transportation systems. To address this issue, \cite{li2018multi} proposed a deep multi-task representation learning model for arrival time estimation. This model produces meaningful representation that preserves various trip properties and at the same time leverages the underlying road network and the spatiotemporal prior knowledge.

 \textbf{Trajectories} \cite{wang2018will,zhang2018deeptravel} tried to estimate the travel time of a path from the mobility trajectory data. \cite{zhang2018deeptravel} proposed a RNN based deep model named DEEPTRAVEL which can learn from the historical trajectories to estimate the travel time. \cite{wang2018will} proposed an end-to-end Deep learning framework for Travel Time Estimation called DeepTTE that estimated the travel time of the whole path directly rather than first estimating the travel times of individual road segments or sub-paths and then summing up them. \cite{martin2018graph} studied the problem of inferring the purpose of a user’s visit at a certain location from trajectory data. They proposed a graph convolutional neural networks (GCNs) for the inference of activity types (i.e., trip purposes) from GPS trajectory data generated by personal smartphones. The mobility graphs of a user is constructed based on all his/her activity areas and connectivities based on the trajectory data, and then the spatio-temporal activity graphs are fed into GCNs for activity types inference. \cite{gao2017identifying} studied the problem of Trajectory-User Linking (TUL), which aims to identify and link trajectories to users who generate them in the LBSNs. A Recurrent Neural Networks (RNN) based model called TULER is proposed to address the TUL problem by combining the check-in trajectory embedding model and stacked LSTM. Identifying the distribution of users’ transportation modes, e.g. bike, train, walk etc., is an essential part of travel demand analysis and transportation planning \cite{dabiri2018inferring,wang2017detecting}. \cite{dabiri2018inferring} proposed a CNN model to infer travel modes based on only raw GPS trajectories, where the modes are labeled as walk, bike, bus, driving, and train.
  
\subsection{Anomaly Detection}
Anomaly detection or outlier detection aims to identify the rare items, events or observations which raise suspicions by differing significantly from the majority of the data. Current works on anomaly detection for ST data mainly focus on the data types of events and spatial maps.

\textbf{Events.} \cite{sun2017dxnat} tried to detect the non-recurring traffic congestions caused by temporal disruptions such as accidents, sports games, adverse weathers, etc. A convolutional neural network (CNN) is proposed to identify non-recurring traffic anomalies that are caused by events. \cite{zhang2018deep} studied how to detect traffic accidents from social media data. They first thoroughly investigated the 1-year over 3 million tweet contents in Northern Virginia and New York City, and then two deep learning methods: Deep Belief Network (DBN) and Long Short-Term Memory (LSTM) were implemented to identify the traffic accident related tweets. \cite{zhu2018deep} proposed to utilize Convolutional Neural Networks (CNN) for automatic detection of traffic incidents in urban networks by using traffic flow data. \cite{chen2016learning} collected big and heterogeneous data including human mobility data and traffic accident data to understand how human mobility will affect traffic accident risk. A deep model of Stack denoise Autoencoder was proposed to learn hierarchical feature representation of human mobility, and these features were used for efficient prediction of traffic accident risk level. 
 
\textbf{Spatial maps.} \cite{liu2016application} presented the first application of Deep Learning techniques as alternative methodology for climate extreme events detection such as hurricanes and heat waves. The model was trained to classify tropical cyclone, weather front and atmospheric river with the climate image data as the input. \cite{kim2017resolution} studied how to detect and localize extreme climate events in very coarse climate data. The proposed framework is based on two deep neural network models, (1) Convolutional Neural Networks (CNNs) to detect and localize extreme climate events, and (2) Pixel recursive recursive super resolution model to reconstruct high resolution climate data from low resolution climate data. To address the issue of limited labeled extreme climate events, \cite{racah2017extremeweather} presented a multichannel spatiotemporal CNN architecture for semi-supervised bounding box prediction. The approach proposed in \cite{racah2017extremeweather} is able to leverage temporal information and unlabeled data to improve the localization of extreme weathers. 

\subsection{Other tasks.} Besides the problems we discussed above, deep learning models are also applied in other STDM tasks include recommendation \cite{chang2018content,li2018next,zhao2017geo}, pattern mining \cite{ouyang2016deepspace},  relation mining \cite{zhou2018trajectory}, etc. \cite{chang2018content} proposed a content-aware hierarchical POI embedding model CAPE for POI recommendation. From text contents, CAPE captures not only the geographical influence of POIs, but also the characteristics of POIs. \cite{zhao2017geo} also proposed to exploit the embedding learning technique to capture the contextual check-in information for POI recommendation. \cite{ouyang2016deepspace} proposed a deep-structure model called DeepSpace to mine the human mobility patterns through analyzing the mobile data of human trajectories. \cite{zhou2018trajectory} studied the problem of Trajectory-User Linking (TUL), which aims to link trajectories to users who generate them from the geo-tagged social media data. A semi-supervised trajectory-user relation learning framework called TULVAE (TUL via Variational AutoEncoder) is proposed to learn the human mobility in a neural generative architecture with stochastic latent variables that span hidden states in RNN.

\begin{figure}[!t]
\begin{center}
\includegraphics[height=4.5cm, angle=0]{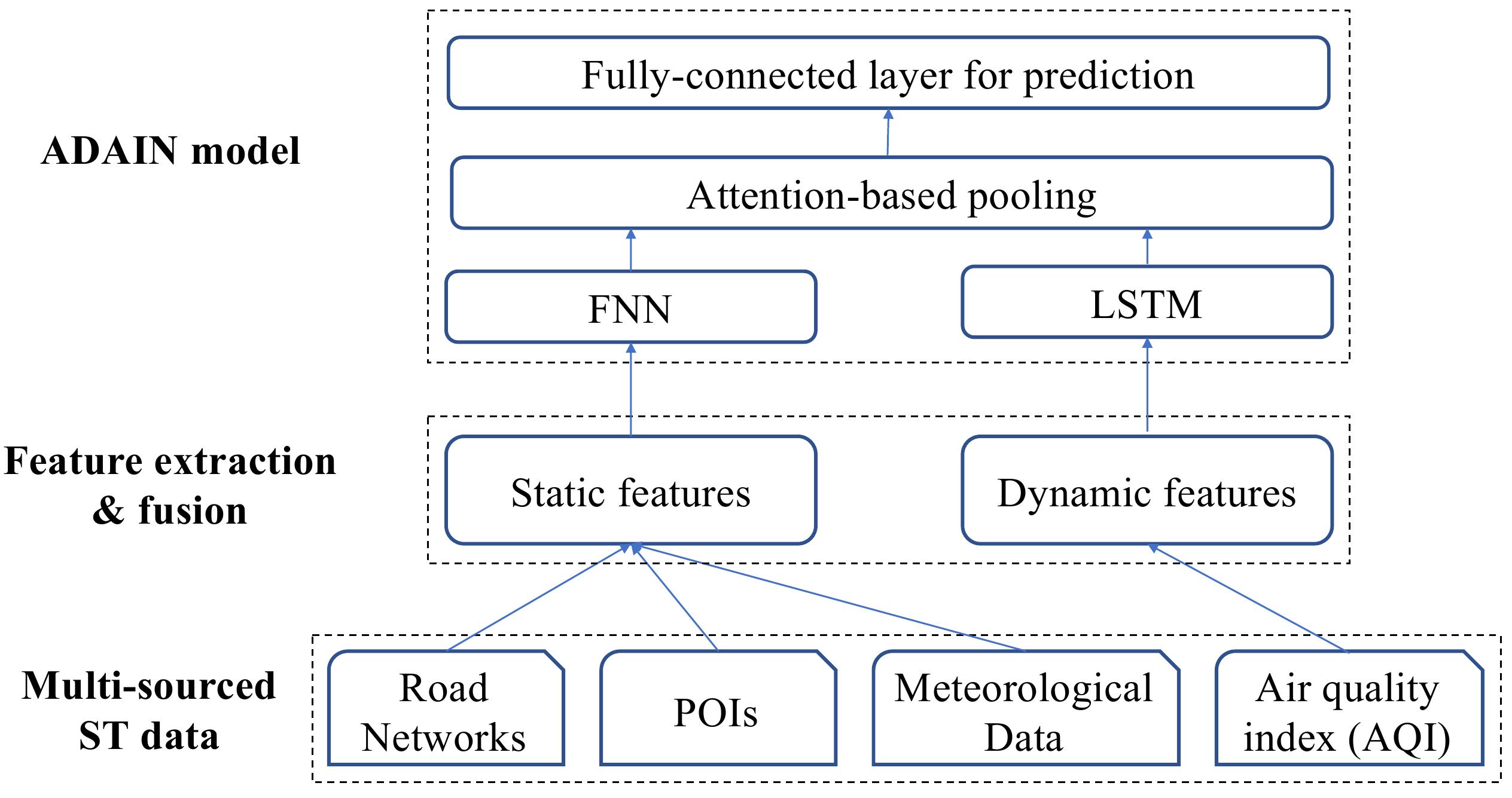}
\caption{ADAIN model framework.}
\label{adain}
\end{center}
\end{figure}
\subsection{Fusing Multi-Sourced Data}
Besides the ST data under study, there are usually some other types of data that are highly correlated to the ST data. Fusing such data together with the ST data can usually improve the performance of various STDM tasks.  For example, the urban traffic flow data can be significantly affected by some external factors such as weather, social events, and holidays. Some recent works try to fuse the ST data and other types of data into a deep learning architecture for jointly learning features and capturing the correlations among them \cite{chen2016learning,cheng2018neural,liao2018dest,yao2018deep,rong2018parking,zhang2017application,yuan2018hetero}. Generally, there are two popular ways to fuse the multi-sourced data in applying deep learning models for STDM, raw data-level fusion and latent feature-level fusion.

\textbf{Raw data-level fusion.} For the raw data-level fusion, the multi-sourced data are integrated first and then input into the deep learning model for feature learning. \cite{yuan2018hetero} studied the traffic accident prediction problem by using the Convolutional Long Short-Term Memory (ConvLSTM) neural network model. First, the entire studied area is partitioned into grid cells. Then a number of fine-grained urban and environmental features such as traffic volume, road condition, rainfall, temperature, and satellite images are collected and map-matched with each grid cell. Given the number of accidents as well as the external features at each location mentioned above as the model input, a Hetero-ConvLSTM model to predict the number of accidents that will occur in each grid cell in future time slots is proposed. \cite{cheng2018neural} proposed the ADAIN model which fused both the urban air quality information from monitoring stations and the urban data that are closely related to air quality, including POIs, road networks and meteorology for inferring fine-grained urban air quality of a city. The framework of ADAIN model is shown in Fig \ref{adain}. Features are first manually extracted from the multi-sourced data including road networks, POIs, meteorological data and urban air quality index data. Then all the features are fused together and then fed into FNN and RNN models for feature learning.

\begin{figure}[!t]
\begin{center}
\includegraphics[height=7.5cm, angle=0]{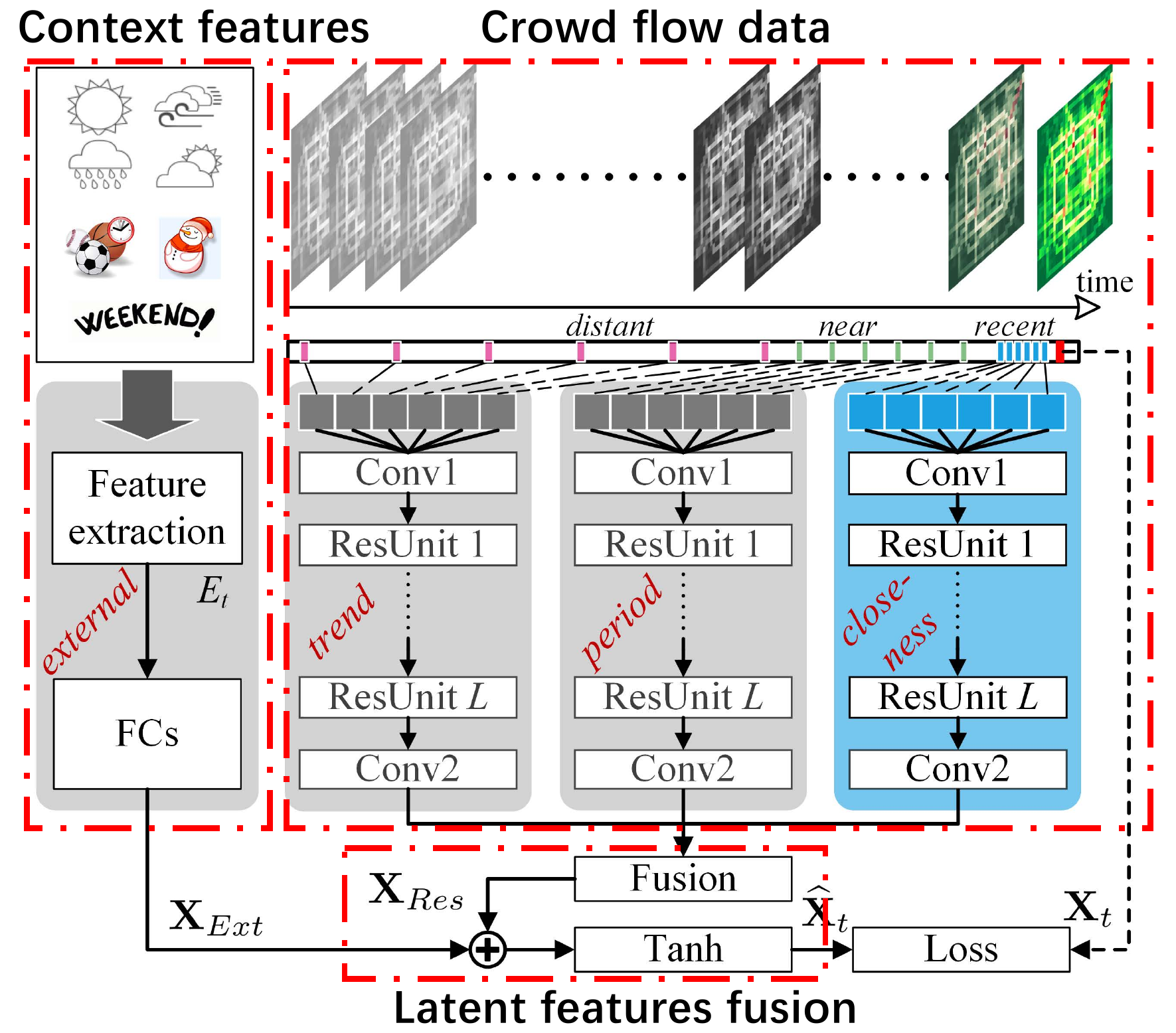}
\caption{ST-ResNet architecture \cite{liao2018dest}.}
\label{ST-ResNet}
\end{center}
\end{figure}

\textbf{Latent feature-level fusion.} For the latent feature-level fusion, different types of raw features are input into different deep learning models first, and then a latent feature fusion component is used to fuse different types of latent features. \cite{liao2018dest} proposed a deep-learning-based approach called ST-ResNet, which is based on the residual neural network framework to collectively forecast the inflow and outflow of crowds in each region of a city. As shown in Fig \ref{ST-ResNet}, ST-ResNet handles two types of data, the ST crowd flow data sequences in a city and the external features including the weather and holiday events. Two components are designed to learn the latent features of the external features and the crowd flow data features seperately, and then a feature fusing function $tanh$ is used to integrate the two types of learned latent features. \cite{yao2018deep} proposed a Deep Multi-View Spatial-Temporal Network (DMVST-Net) framework to combine multi-view data for taxi demand prediction. DMVST-Net consists of three views: temporal view, spatial view and semantic view. CNN is used to learn features from the spatial view, LSTM is used to learn features from the temporal view and network embedding is applied to learn the correlations among regions. Finally, a fully connected neural network is applied to fuse all the latent features of the three views for taxi demand prediction. 

\subsection{Attention Mechanism}
Attention is a mechanism that was developed to improve the performance of the Encoder-Decoder RNN on machine translation \cite{attention}.   A major limitation of the Encoder-Decoder RNN is that it encodes the input sequence to a fixed length internal representation, which results in worse performance for long input sequences. To address this issue, attention allows the model to learn which encoded words in the source sequence to pay attention to and to what degree during the prediction of each word in the target sequence. Although attention is initially proposed in machine translation with the word sequence data as the input, it actually can be applied to any kind of inputs such as images, which is called visual attention. As many ST data can be represented as sequential data (time series and trajectories) and image-like spatial maps, attention can also be incorporated into deep learning model for improving the performance of various STDM tasks \cite{cheng2018neural,feng2018deepmove,fernando2018soft+,huang2018deepcrime,li2018next,liang2018geoman,liu2018attentive,varshneya2017human,zhou2018predicting}. 

The neural attention mechanism used in STDM can be generally categorized into spatial domain attention \cite{cheng2018neural,fernando2018soft+} and temporal domain attention \cite{feng2018deepmove,huang2018deepcrime,li2018next,zhou2018predicting}. Some works use both spatial and temporal domain attentions \cite{liang2018geoman,liu2018attentive,varshneya2017human}. \cite{fernando2018soft+} proposed a combined attention model in the spatial domain. It utilizes both ``soft attention'' as well as ``hard-wired'' attention in order to map the trajectory information from the local neighborhood to the future positions of the pedestrian of interest. \cite{huang2018deepcrime} proposed an attentive hierarchical recurrent network model named DeepCrime for crime prediction. The temporal domain attention mechanism is applied to capture the relevance of crime patterns learned from previous time slots in assisting the prediction of future crime occurrences, and automatically assign the importance weights to the learned hidden states at different time frames. In the proposed attention mechanism, the importance of crime occurrence in the past time slots is estimated by deriving a normalized importance weight via a softmax function. \cite{liang2018geoman} proposed a multi-level attention network for predicting the geo-sensory time series that are generated by sensors deployed in different geospatial locations to continuously and cooperatively monitor the surrounding environment, such as air quality. Specifically, in the first level attention, a spatial attention mechanism consisting of local spatial attention and global spatial attention is proposed to capture the complex spatial correlations between different sensors’ time series. In the second level attention, a temporal attention is applied to model the dynamic temporal correlations between different time intervals in a time series. 

\section{Applications}
Large volumes of ST data are generated from various application domains such as transportation, on-demand service, climate \& weather, human mobility, location-based social network (LBSN), crime analysis, and neuroscience. Table \ref{app} shows the related works of the application domains mentioned above. One can see that the largest proportion of the works fall into transportation and human mobility due to the increasing availability of the urban traffic data and human mobility data. In this section, we will describe the applications of deep learning techniques used for STDM in different applications.
\begin{table*}
\caption{Related works in different application domains}
\begin{center}
\begin{tabular}{|c|p{12cm}|}
\hline
Application domains & Related works\\
\hline
Transportation & \cite{zhang2018deep,zhu2018deep,ren2018deep,duan2014deep,du2018hybrid,bao2019spatiotemporal,duan2016efficient,soua2016big,huang2014deep,polson2017deep,Dixon2017Deep,nguyen2018deep,yu2017deep,liao2018deep,cui2016deep,kim2017deeprain,niu2014deepsense,song2016deeptransport,liao2018dest,wang2017detecting,li2018diffusion,sun2017dxnat,wang2018efficient,wang2018enhancing,kurth2018exascale,ma2018forecasting,yuan2018hetero,cui2018high,dabiri2018inferring,ma2015large,lv2018lc,chen2016learning,ma2017learning,lin2018pattern,chen2018pcnn,yuspatio,yu2017spatiotemporal,lv2015traffic,wang2016traffic,liu2018urban,MDM19STCNN,MDM19TGCN,AAAI19Traffic,AAAIAGCN,AAAI19DGCN} \\
\hline
On-demand Service & \cite{ai2018deep,chai2018bike,rodrigues2019combining,yao2018deep,wang2017deepsd,lee2018forecasting,ke2018hexagon,li2018origin,zhou2018predicting,ke2017short,geng2019spatiotemporal,zhao2018temporal,WWW19MoD,AAAI19MGC,AAAI19Bike}\\
\hline
Climate \& Weather & \cite{cheng2018neural,chattopadhyay2018test,liu2016application,zhang2017application,xingjian2015convolutional,tao2016deep,cheng2018ensemble,kurth2018exascale,lin2018exploiting,giffard2018fused,ke2017short,zhang2018short,liu2018smart,liu2018smart1,scher2018toward,chen2017using,roesch2017visualization,zhu2018wind} \\
\hline
Human Mobility & \cite{karatzoglou2018convolutional,yang2017neural,wang2016action,wang2015action,liu2018attentive,endo2016classifying,xu2018collision,DBLP,li2018deep,jiang2018deep,zhang2017deep,feng2018deepmove,ouyang2016deepspace,song2016deeptransport,zhang2018deeptravel,nguyen2012extracting,zheng2016generating,martin2018graph,varshneya2017human,gao2017identifying,yao2018learning,wang2018learning,wu2017modeling,li2018multi,zhao2018prediction,fan2018online,zonoozi2018periodic,zhang2018predicting,endo2017predicting,liu2016predicting,chow2018representation,yao2017serm,zhang2018short,gupta2018social,alahi2016social,shen2018stepdeep,lv2018t,zhao2018temporal,yao2017trajectory,gao2018trajectory,jiang2017trajectorynet,zhou2018trajectory,chen2018trip2vec,wang2018will,MDM19POI,AAAI19DSTN} \\
\hline
Location Based Social Network & \cite{yang2017neural,chang2018content,ding2018geographical,zhao2017geo,kong2018hst,li2018next,liu2016predicting,yang2018recurrent,yang2018spatio,zhou2018trajectory,yang2018unsupervised,zhao2018go}\\
\hline
Crime Analysis & \cite{duan2017deep,shi2017deep,wang2017deep1,huang2018deepcrime,racah2017extremeweather,hossain2015forecasting,rasp2018neural,kim2017resolution,zaytar2016sequence}\\
\hline
Neuroscience & \cite{plis2014deep,wen2018deep,marblestone2016toward,dvornek2017identifying,guo2017diagnosing,heinsfeld2018identification,huang2016latent,huang2018modeling,jang2017task,jin2014classification,kim2016deep,meszlenyi2017resting,nie20163d,sarraf2016deep,horikawa2017generic,kleesiek2016deep,kawahara2017brainnetcnn,shi2018multimodal,korolev2017residual,zhang2016identifying}\\
\hline
\end{tabular}
\end{center}
\label{app}
\end{table*}%

\subsection{Transportation}
With the increasing availability of transportation data collected from various sensors like loop detector, road camera, and GPS, there is an urgent need to utilize deep learning methods to learn the complex and highly non-linear spatio-temporal correlations among the traffic data to facilitate various tasks such as traffic flow prediction \cite{du2018hybrid,huang2014deep,liao2018deep,polson2017deep,soua2016big,yang2017optimized}, traffic incident detection \cite{ren2018deep,zhang2018deep,zhu2018deep} and traffic congestion prediction \cite{ma2015large,sun2017dxnat}. Such transportation related ST data usually contain information of the traffic speed, volume, or traffic incidents, the locations of the road segments of regions, and the time. Transportation data can be modeled as time series, spatial maps and ST raster in different application scenarios. For example, in road network-scale traffic flow prediction the traffic flow data collected from multiple road loop sensors can be modeled as a raster matrix where one dimension is the locations of the sensors and the other is the time slots \cite{ma2017learning}. The loop sensors can be also connected as a sensor graph based on the connections among the road links where the sensors are deployed, and the the traffic data of a road network can be modeled as a graph spatial map so that GraphCNN models can be applied \cite{li2018diffusion,yuspatio}. While in road-level traffic prediction, the historical traffic flow data on each single road is modeled as a time series, and then RNN or other deep learning models are used for traffic prediction of a single road \cite{huang2014deep,liao2018deep,yang2017optimized}. 

\subsection{On-Demand Service}
In recent years, various on-demand services such as Uber, Mobike, DiDi, GoGoVan have become increasingly popular due to the wide use of mobile phones. The on-demand services have taken over the traditional businesses by serving people with what and where they want. Many on-demand services produce a large number of ST data which involve the locations of the customers and the required service time. For example, Uber and DiDi are two popular ride-sharing on-demand service providers in USA and China, respectively. They both provide services including taxi hailing, private car hailing, and social ride-sharing to users via a smartphone application. To better meet customers' demand and improve the service, a crucial problem is how to accurately predict the demand and supply of the service at different locations and time. Deep learning methods for STDM in the application of on-demand service mostly focus on predicting the demand and supply. \cite{ai2018deep} proposed to apply deep learning methods to forecast the demand-supply distributions of the dockless bike-sharing system. \cite{lin2018predicting} proposed a graph CNN model to predict the station-level hourly demand in a large-scale bike-sharing network. \cite{rodrigues2019combining,yao2018deep} proposed to use LSTM model to predict the taxi demand in different areas. \cite{wang2017deepsd} applied ResNet model to predict the supply-demand for online car-hailing services. The historical demand-supply in different regions of the city under study is usually modeled as spatial maps or raster tensors, so that CNN, RNN and combined models are applied to predict the future.

\subsection{Climate \& Weather}
Climate science is the scientific study of climate, scientifically defined as weather conditions averaged over a period of time. The weather data usually contain the atmospheric and oceanic conditions (e.g., temperature, pressure, wind-flow, and humidity) that are collected by various climate sensors deployed at fixed or floating locations. As the climate data of different locations usually present high spatio-temporal correlations, STDM techniques are widely used for short-term and long-term weather forecasting. Especially, with the recent advances of deep learning techniques, many works tried to incorporate deep learning models for analyzing various weather and environment data \cite{kurth2018exascale,scher2018toward}, such as air quality inference \cite{cheng2018neural,lin2018exploiting}, precipitation prediction \cite{liu2016application,xingjian2015convolutional}, wind speed prediction \cite{liu2018smart,zhu2018wind}, and extreme weather detection \cite{liu2016application}. The data related to climate and weather can be spatial maps (e.g. radar reflectivity images) \cite{zhang2017application}, time series (e.g. wind speed) \cite{cheng2018ensemble}, and events (e.g. extreme weather events) \cite{liu2016application}. \cite{cheng2018neural} proposed a neural attention model to predict the urban air quality data of different monitoring stations. \cite{liu2016application} proposed to use CNN model for detecting extreme weather in climate databases. CNN model can be also used to estimate the precipitation from the remote sensing images \cite{liu2016application}.

\subsection{Human Mobility}
With the wide use of mobile devices, recent years have witnessed an explosion of extensive geolocated datasets related to human mobility. The large volume of human mobility data enable us to quantitatively study individual and collective human mobility patterns, and to generate models that can capture and reproduce the spatiotemporal structures and regularities in human trajectories. The study of human mobility is especially important for applications such as estimating migratory flows, traffic forecasting, urban planning, human behavior analysis, and personalized recommendation. Deep learning techniques applied on human mobility data mostly focus on human trajectory data mining such as trajectory classification \cite{endo2016classifying}, trajectory prediction \cite{feng2018deepmove,jiang2018deep,xu2018collision}, trajectory representation learning \cite{li2018deep,yao2018learning}, mobility pattern mining \cite{ouyang2016deepspace}, and human transportation mode inference from trajectories \cite{dabiri2018inferring,gao2017identifying}. Based on different application scenarios and analytic purposes, human trajectories can be modeled as different types of ST data types and data representation so that different deep learning models can be applied. The most widely used models for human trajectory data mining are RNN an CNN models, and sometimes the two types of models are combined to capture both the spatial and temporal correlations among the human mobility data.
\subsection{Location Based Social Network (LBSN)}
Location-based social networks such as Foursquare and Flickr are social networks that use GPS features to locate the users and let the users broadcast their locations and other contents from their mobile device \cite{LBSN}. A LBSN does not only mean adding a location to an existing social network so that people can share location-embedded information, but also consists of the new social structure made up of individuals connected by their locations in the physical world as well as their location-tagged media content. LBSN data contain a large number of user check-in data which consists of the instant location of an individual at a given timestamp. Currently, deep learning methods have been used in analyzing the user generated ST data in LBSN, and the studied tasks include next check-in location prediction \cite{karatzoglou2018convolutional}, user representation learning in LBSN \cite{yang2017neural}, geographical feature extraction \cite{ding2018geographical} and user check-in time prediction \cite{yang2018recurrent}.  

\subsection{Crime Analysis}
Law enforcement agencies store information about reported crimes in many cities and make the crime data publicly available for research purposes. The crime event data typically has the type of crime (e.g., arson, assault, burglary, robbery, theft, and vandalism), as well as the time and location of the crime. Patterns in crime and the effect of law enforcement policies on the amount of crime in a region can be studied using this data with the goal of reducing crime \cite{atluri2018spatio}. As the crimes happened at different regions of a city usually present high spatial and temporal correlations, deep learning models can be used with the crime account heat map of a city as the input to capture such complex correlations \cite{duan2017deep,huang2018deepcrime,wang2017deep1}. For example, \cite{duan2017deep} proposed a Spatiotemporal Crime Network based on CNN to forecast the crime risk of each region in the urban area for the next day. \cite{wang2017deep1} proposed to utilize ST-ResNet model to collectively predict crime distribution over the Los Angeles area. \cite{huang2018deepcrime} developed a new crime prediction framework--DeepCrime, which is a deep neural network architecture that uncovers dynamic crime patterns and explores the evolving inter-dependencies between crimes and other ubiquitous data in urban space. As we discussed before, crime data are typical ST event data, but are usually represented as spatial maps through merging the data in spatial and temporal domains so that deep learning models can be applied for analytics.

\subsection{Neuroscience}

In recent years, brain imaging technology has become a hot topic within the field of neuroscience. Such technology includes functional Magnetic Resonance Imaging (fMRI), electroencephalography (EEG), Magnetoencephalography (MEG), and functional Near Infrared Spectroscopy (fNIRS). The spatial and temporal resolutions of neural activity measured by these technologies is quite different from another. fMRI measures the neural activity from millions of locations, while it is only measured from tens of locations for EEG data. fMRI typically measures activity for every two seconds, while the temporal resolution of EEG data is is typically 1 millisecond. Because of its space resolving power, fMRI and EEG combined with deep learning methods, has been widely used in the study of neuroscience \cite{dvornek2017identifying,jang2017task,meszlenyi2017resting,sarraf2016deep}. 
As we discussed before, deep learning models are mostly used for the classification task in neuroscience by using the fMRI data or EEG data such as disease classification \cite{dvornek2017identifying}, brain function network classification \cite{meszlenyi2017resting} and brain activation classification \cite{jang2017task}. For example, Long-Short Term Memory network (LSTM) was used to identify Autism Spectrum Disorder (ASD) \cite{dvornek2017identifying}, Convolutional Neural Networks (CNN) were used to diagnose amnestic Mild Cognitive Impairment (aMCI) \cite{meszlenyi2017resting} and Feedforward Neural Networks (FNN) were used to classify Schizophrenia \cite{ICVGIP16}.


\section{Open Problems}
Though many deep learning methods have been proposed and widely used for STDM in diverse application domains discussed above, challenges still exist due to the highly complex, large volume, and fast increasing ST data. In this section, we provide some open problems that have not been well addressed by  current works and need further studies in the future.

\textbf{Interpretable models.} Current deep learning models for STDM are mostly considered as \emph{black-boxs} which lack of interpretability. Interpretability gives deep learning models the ability to explain or to present the model behaviors in understandable terms to humans, and it is an indispensable part for machine learning models in order to better serve people and bring benefits to society \cite{Interpretable}. Considering the complex data types and representations of ST data, it is more challenging to design interpretable deep learning models compared with other types of data such as images and word tokens. Although attention mechanisms are used in some previous works to increase the model interpretability such as periodicity and local spatial dependency \cite{cheng2018neural,huang2018deepcrime,liang2018geoman}, how to build a more interpretable deep learning model for STDM tasks is still not well studied and remains an open problem.

\textbf{Deep learning model selection.} For a given STDM task, sometimes multiple types of related ST data can be collected and different data representations can be choosen. How to properly select the ST data representations and the corresponding deep learning modes is not well studied. For example, in traffic flow prediction, some works model the traffic flow data of each road as a time series so that RNN, DNN or SAE are used for prediction \cite{lv2015traffic,soua2016big}; some works model the traffic flow data of multiple road links as spatial maps so that CNN is applied for prediction \cite{zhang2016dnn}; and some works model the traffic flow data of a road network as a graph so that GraphCNN is adopted \cite{li2018diffusion}. There lacks of deeper studies on how to properly select deep learning models and data representations of the ST data for better addressing the STDM task under study.

\textbf{Broader applications to more STDM tasks.} Although deep learning models have been widely used in various STDM tasks discussed above, there are some tasks that have not been addressed by deep learning models such as frequent pattern mining and relationship mining  \cite{atluri2018spatio,pattern}. The major advantage of deep learning is its powerful feature learning ability, which is essential to some STDM tasks such as predictive learning and classification that largely rely on high quality features. However, for some STDM tasks like frequent pattern mining and relationship mining, learning high quality features may not be that helpful because these tasks do not require features. Based on our review, currently there are very few or even no works that utilize deep learning models for the tasks mentioned above. So it remains an open problem that how deep learning models along or the integration of deep learning models with traditional models such as frequent pattern mining and graphical models can be extented to broader applications to more STDM tasks.

\textbf{Fusing multi-modal ST datasets.} In big data era, multi-modal ST datasets are increasingly available in many domains such as neuroimaging, climate science and urban transportation. For example, in neuroimaging, fMRI and DTI can both capture the imaging data of the brain activity with different technologies that provide different spatiotemporal resolutions \cite{fMRI}. How to use deep learning models to effectively fuse them together to better perform the tasks of disease classification and brain activity recognition is less studied. The multi-modal transportation data including taxi trajectory data, bike-sharing trip data and public transport check-in/out data of a city can all reflect the mobility of urban crowd flow from different perspectives \cite{du2018hybrid}. Fusing and analyzing them together rather than separately can more comprehensively capture the underlying mobility patterns and make more accurate predictions. Although there are recent attempts that tried to apply deep learning models for transferring knowledge from the crowd flow data among different cities \cite{DBLP,WWW19City}, how to fuse multi-modal ST datasets with deep learning models is still not well studied and needs more research attention in the future.

\section{Conclusion}
In this paper, we conduct a comprehensive overview of recent advances in exploring deep learning techniques for STDM. We first categorize the different data types and representations of ST data and briefly introduce the popular deep learning models used to STDM. For different types of ST data and their representations, we show the corresponding deep learning models that are suitable to handle them. Then we give a general framework showing the pipeline of utilizing deep learning models for addressing STDM tasks. Under the framework, we overview current works based on the categorization of the ST data types and the STDM tasks including predictive learning, representation learning, classification, estimation and inference, anomaly detection, and others. Next we summarize the applications of deep learning techniques for STDM in different domains including transportation, on-demand service, climate \& weather, human mobility, location-based social network (LBSN), crime analysis, and neuroscience. Finally, we list some open problems and point out the future research directions for this fast growing research filed.



\bibliographystyle{abbrv}
\bibliography{DL2.bib}
\end{document}